\documentclass[nohyperref]{article}

\usepackage{todonotes, soul}

\usepackage{arxiv}

\usepackage[utf8]{inputenc} 
\usepackage[T1]{fontenc}    
\usepackage{hyperref}       
\usepackage{url}            
\usepackage{booktabs}       
\usepackage{amsfonts}       
\usepackage{nicefrac}       
\usepackage{microtype}      
\usepackage{lipsum}		
\usepackage{graphicx}
\usepackage{natbib}
\usepackage{doi}

\usepackage{graphicx}
\usepackage{subcaption}
\usepackage{placeins}

\usepackage{blkarray}      
\usepackage{multirow}
\usepackage{caption} 
\usepackage{array}
\newcommand{\PreserveBackslash}[1]{\let\temp=\\#1\let\\=\temp}
\newcolumntype{C}[1]{>{\PreserveBackslash\centering}p{#1}}

\usepackage{amsmath}
\usepackage{amssymb}
\usepackage{mathtools}
\usepackage{amsthm}
\usepackage{thmtools, thm-restate}

\usepackage{algorithmic}
\usepackage{algorithm}

\usepackage{pifont}
\newcommand{\xmark}{\ding{56}}

\DeclareMathOperator*\argmin{arg\,min}
\DeclareMathOperator*\argmax{arg\,max}

\theoremstyle{plain}

\theoremstyle{definition}

\newcommand{\Normal}{\mathcal{N}}

\title{Combining Multi-Fidelity Modelling and Asynchronous Batch Bayesian Optimization}

\date{} 					

\author{ Jose Pablo Folch \thanks{Corresponding author.} \\
	Imperial College London\\
	London, UK \\
	\texttt{jose.folch16@imperial.ac.uk} \\
	\And
	Robert M Lee \\
	BASF SE \\
	Ludwigshafen, Germany \\
	\texttt{robert-matthew.lee@basf.com} \\
	\AND
	Behrang Shafei \\
	BASF SE \\
	Ludwigshafen, Germany \\
	\texttt{behrang.shafei@basf.com} \\
	\And
	David Walz \\
	BASF SE \\
	Ludwigshafen, Germany \\
	\texttt{david-simon.walz@basf.com} \\
	\And
	Calvin Tsay \\
	Imperial College London\\
	London, UK \\
	\texttt{c.tsay@imperial.ac.uk} \\
	\AND
	Mark van der Wilk \\
	Imperial College London\\
	London, UK \\
	\texttt{m.vdwilk@imperial.ac.uk} \\
	\And
	Ruth Misener \\
	Imperial College London\\
	London, UK \\
	\texttt{r.misener@imperial.ac.uk} \\
}

\renewcommand{\shorttitle}{Combining Multi-Fidelity Modelling and Asynchronous BBO}

\begin{document}
\maketitle

\begin{abstract}
Bayesian Optimization is a useful tool for experiment design. Unfortunately, the classical, sequential setting of Bayesian Optimization does not translate well into laboratory experiments, for instance battery design, where measurements may come from different sources and their evaluations may require significant waiting times. Multi-fidelity Bayesian Optimization addresses the setting with measurements from different sources. Asynchronous batch Bayesian Optimization provides a framework to select new experiments before the results of the prior experiments are revealed.
This paper proposes an algorithm combining multi-fidelity and asynchronous batch methods. We empirically study the algorithm behavior, and show it can outperform single-fidelity batch methods and multi-fidelity sequential methods. As an application, we consider designing electrode materials for optimal performance in pouch cells using experiments with coin cells to approximate battery performance.
\end{abstract}

\keywords{Bayesian Optimization \and Machine Learning \and Batch Optimization \and Asynchronous \and Multi-fidelity}

\section{Introduction} \label{sec: intro}

\begin{figure}[ht]
	\centering
	\includegraphics[width = 0.6\textwidth]{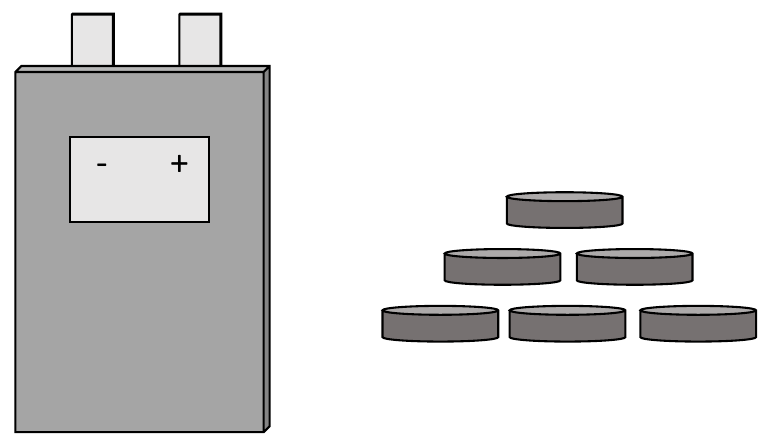}
	\caption{Motivating example. On the left is pouch-cell battery and on the right are coin cell batteries \citep{CHEN20191094,DORFLER2020539,liu2021strategy}. Using coin cells, which are cheaper and faster to make than pouch batteries, we can approximate the behaviour of our target. Usually these experiments are carried out in large batches.}
	\label{fig: battery_illustration}
\end{figure}

The optimal design of many engineering processes can be subject to expensive and time-consuming experimentation. For efficiency, we seek to avoid wasting valuable resources in testing sub-optimal designs. One way to achieve this is by obtaining cheaper approximations of the desired system, which allow us to quickly explore new regimes and avoid areas that are clearly sub-optimal. As an example, consider the case diagrammed in Figure \ref{fig: battery_illustration} from battery materials research with the goal of designing electrode materials for optimal performance in pouch cells. We can use experiments with cheaper coin cells and shorter test procedures to approximate the behaviour of the material in longer stability tests in pouch cells, which is in turn closer to the expected performance in electric car applications \citep{CHEN20191094,DORFLER2020539,liu2021strategy}. Similarly, design goals regarding battery life such as discharge capacity retention can be approximated using an early prediction model on the first few charge cycles rather than running aging and stability tests to completion \citep{attia2020closed}.

Furthermore, in most applications it is infeasible to wait for a single experiment to end before planning new ones, due to the long waiting times for the experiments to finish (from a few weeks to months). This means that in practice experiments will be carried out in batch or in parallel. However, owing to the introduction of cheaper approximations and variations in material degradation rates, experiments may not finish simultaneously. For example, if our experimental batch consists of a mixture of cells tested with slower and accelerated test procedures, some results will become available in a few weeks, whereas others can take \textit{months}. In the meantime, we do not want to leave resources idle while we wait for all experiments to be finished. Therefore, we must select new experiments before receiving the full results from those still in progress. This inspires the question: \textit{How can we automatically and continuously design new experiments when the prior experimental data returns on different (and uncertain) time-frames and the type of measurements may vary?}

Design of experiments is a well-studied area, with applications ranging from parameter estimation \citep{asprey2000statistical, box1959design,waldron2019autonomous} and model selection \citep{hunter1965designs, tsay2017doeforparameterestimation} in chemistry, to psychology survey design \citep{vincent2016hierarchical, foster2021deep}. Many of these works assume an underlying model which we want to learn efficiently. 
See \citet{franceschini2008model} for a comprehensive overview of model-based design of experiments and \citet{wang2022pyomo} for computational considerations. 
However, our particular focus lies in designing experiments for optimizing black box functions \citep{Bajaj2021blackbox}. Specifically, these are complex and expensive-to-evaluate functions from which we do not receive gradient information. We can select a sequence of function inputs, from which we observe a sequence of outputs.

Bayesian Optimization (BO) \citep{jones1998efficient, shahriari2016bo} optimizes black-box functions using probabilistic surrogate models, which manage the trade-off between exploiting promising experiments and exploring unseen design space \citep{bhosekar2018advances}. Such methods have been successfully applied in the area of chemical engineering \citep{thebelt2022maximizingcheminfo}.

This paper focuses on recent advances in multi-fidelity BO \citep{kandasamy2019multi, takeno2020multi} and Batch BO \citep{gonzalez2016batch, alvi2019localpen, pmlr-v84-kandasamy18a, snoek2012practical}. Multi-fidelity BO uses cheap approximations of the objective function to speed up the optimization process. Batch BO selects multiple experiments at the same time. We explore the close relationship between both research areas, provide an overview of the current state-of-the-art, and introduce a general algorithm that allows practitioners to \textit{simultaneously} take advantage of multi-fidelity and batching while using any acquisition function of their choice, significantly speeding up Bayesian Optimization procedures. We show the usefulness of the method by applying it to synthetic benchmarks, a high-dimensional example, and real life-experiments relating to coin and pouch cell batteries tested with different measurement procedures.

The contributions of the paper include:
\begin{itemize}
    \item[I.] A brief review of state-of-the-art multi-fidelity and batch Bayesian optimization methods.
    \item[II.] We propose an algorithm that allows us to incorporate multi-fidelity and batch Bayesian optimization to any acquisition function. Allowing many acquisition functions extends the current literature, because previous contributions are limited to a  single acquisition function each.
    \item[III.] We provide an empirical study, that showcases the benefit of adding multi-fidelity and parallelization to black-box optimization.
\end{itemize}

\section{Related Work}

Bayesian Optimization (BO) has been well researched in the domain of tuning the hyper-parameters of computationally expensive machine learning models \citep{bergstra2011algorithms,li2017hyperband}. However, to be efficiently applicable to chemical engineering there is a need to adapt and improve such methods \citep{amaran2016simulation,thebelt2022maximizingcheminfo,WANG2022100728}. For example, \cite{folch2022snake} considers the physical cost of changing from one experimental set-up to another and \cite{thebelt2021entmoot,thebelt2022multi,thebelt2022tree} allow input constraints in addition to the black-box functions. The related area of \emph{grey-box optimization} seeks to solve optimal decision-making problems incorporating both black-box, data-driven models and mechanistic, equation-based models \citep{cozad2014learning,boukouvala2017argonaut,olofsson2018bayesian,park2018multi,paulson2022cobalt,kudva2022constrained}. 

The multi-fidelity setting has been gaining attention in machine learning research recently. Multi-fidelity BO uses cheap approximations of the objective function to explore the input space more efficiently, and saves full evaluations for more promising areas. Different approaches for modelling the multi-fidelity using Gaussian Processes (GPs) have been developed \citep{kennedy2000multifidelity,gratiet2013recursive,cutajar2019deep}. In terms of Bayesian Optimization, there has been research from the basic k-armed bandit problem \citep{kandasamy2016multifidelityKbandit} to more general non-GP based methods \citep{sen18a}. This paper addresses the setting and method introduced in \cite{kandasamy2019multi}, where we use an independent GP to model each fidelity. Each fidelity is linked by a bias assumption which allows the transfer of information by taking the tightest of many upper confidence bounds. As a more efficient alternative, we will also explore multi-task Gaussian Processes \citep{journel1976mining,goovaerts1997geostatistics,alvarez2012kernels}, and information-based multi-fidelity~\citep{takeno2020multi}.

Multi-fidelity shares a close relationship with transfer learning \citep{zhuang2021transferlearningsurvery, pan2009survey}, which has seen many successful engineering applications \cite{li2020transfer, li2022conceptual, jia2020transfer, rogers2022transfer}. In transfer learning, we seek to get improved performance in a target task by using data from similar tasks that have been carried out in the past. Multi-fidelity optimization can be thought of as transfer learning, where we can actively \textit{choose} to increase the data of auxiliary data-sets.

\textit{Asynchronous} BO chooses queries while waiting for delayed observations. \cite{pmlr-v84-kandasamy18a} proposed a method based on Thompson Sampling, using the method's natural randomness to ensure variety in experiments. \cite{ginsbourger2011dealing} introduce a variant of Expected Improvement, which seeks to integrate out ongoing queries, and uses Monte Carlo methods to estimate the acquisition function. \cite{snoek2012practical} introduces a more general version of this idea that `fantasizes' delayed observations. \cite{gonzalez2016batch} and \cite{alvi2019localpen} propose mimicking sequential BO by \textit{penalizing} the acquisition function at points where we are asynchronously running experiments.

An example where multi-fidelity approximations are being applied is on the area of molecule design and synthesis \citep{alshehri2020deep}, where we can use cheap computer approximations to search the space of molecules, and then choose which real experiments to carry out based on the best performing simulations \citep{coley2021defining}. \cite{coley2019robotic} use computer simulations to determine feasibility in synthesizing organic compounds, and then use a robotic arm that carries out experiments in batch. This is could be seen as having a multi-fidelity step followed by a batching step, however, the methods are carried out separately.

Recently, \cite{li2021batch} combined batch and multi-fidelity methods using Bayesian Neural Networks (BNNs). This work focuses on simpler, more robust and computationally feasible methods that work directly on the asynchronous setting. \cite{takeno2020multi} propose a multi-fidelity BO algorithm based on max-entropy search \citep{wang2017max}, specializing in asynchronous batching, we take inspiration from this work and generalize it to help avoid the computational complexities of the method, and possible pitfalls from being constrained to a single acquisition function. More recently, GIBBON \citep{moss2021gibbon} was proposed as a general-purpose extension of max-entropy search, by using cheap-approximations to the information gain.

\section{Multi-fidelity Modelling}

Optimal experiment design relies on surrogate models to convey information about our knowledge of the search space. The surrogate model represents our \textit{belief} of how the underlying black-box function looks like. Selecting new experiments near our surrogate's optimum is known as \textit{exploiting}, which is when we choose new experiments close to the best experiments we have observed so far. It is also important for the model to provide some measure of \textit{uncertainty} \citep{hullen2020managing}, which we can leverage to \textit{explore} the search space. Uncertainty should be high in areas where we do not have any data, and low where we have a lot of data.

In particular, this paper focuses on multi-fidelity modelling, where we assume our data-set is formed by a combination of smaller data-sets of varying quality. A few concrete examples where this happens include:
\begin{itemize}
    \item[(a)] The primary motivating example behind this research paper. In battery research, we may conduct experiments on pouch cell batteries, or cheaper experiments using coin cell batteries.
    \item[(b)] \citet{kennedy2000multifidelity} consider the case of combining an expensive computational code which can also be run much faster at a lower level of sophistication, resulting in two data-sets.
    \item[(c)] In machine learning, an expensive model, such as a neural network, can be quickly trained on a subset of the training data to approximate the behaviour of the fully-trained algorithm.
\end{itemize}

We focus on the case where we have a discrete number of fixed fidelities, which in practice will be two or three. We note that while we may be able to choose which fidelity to experiment on, we cannot choose the number of fidelities. For interested readers, \citet{kandasamy2017multicontinous} provide a treatment Bayesian Optimization with a continuous fidelity space. An example of such a fidelity space is early termination of training in a machine learning model where we can choose how long to train for e.g. choose any time between 2 hours and 3 days (in this case the longer we train, the better the approximation). 

Formally, we assume we are trying to model a function $f$. However, we have access to $M$ auxiliary functions, $f^{(m)}$, which approximate $f$ for $m = 1, ..., M$. Each function will have a data-set, $D^{(m)} = \{(x^{(m)}_i, y^{(m)}_i)\}_{i=1}^{N_m}$ consisting of noisy observations of each function:
\begin{equation}
    y^{(m)}_i = f^{(m)}(x^{(m)}_i) + \eta^{(m)} \epsilon_i, \label{eq: fidelity_equation}
\end{equation}
where $\epsilon_i \sim \Normal(0, 1)$. We want a model that will use information from all data-sets to have a more accurate model of the objective $f =: f^{(M)}$. For example, in the battery example, we have access to $M = 2$ types of data -- the cheaper coin cell battery data, and the target pouch cell battery data. We want to use both types of data together, to obtain a more accurate model for pouch cell batteries.

This paper uses Gaussian Process (GP) surrogate models \citep{rasmussen2005gps}, which are very flexible and have well calibrated uncertainty estimates. A GP is a stochastic processes defined by a mean function, $\mu(\cdot): \mathcal{X} \rightarrow \mathbb{R}$, and a positive-definite co-variance function $k(\cdot, \cdot): \mathcal{X} \times \mathcal{X} \rightarrow \mathbb{R}$. In particular, given a data-set $D = \{(x_i, y_i)\}_{i=1}^N$ if we select a GP prior on $f \sim \mathcal{GP}(\mu_0, k_0)$, then we can consider the Gaussian likelihood:
\begin{equation*}
    y \ | \ f, \eta^2 \sim \Normal(f, \ \eta^2 I).
\end{equation*}
This allows us to explicitly compute the posterior of $f$, which is also a GP. More precisely, $f \ | \ D_t \sim \mathcal{GP}(\mu_t,\  k^2_t)$:
\begin{align*}
    \mu_t(x) &= \mu_0(x) + k_0(x)^T( K + \eta^2 I )^{-1} (y - m) \\
    k^2_t(x, x') &= k_0(x, x') - k(x)^T ( K + \eta^2 I ) ^{-1} k(x'),
\end{align*}
where $m_i = \mu_0(x_i)$, $k(x)_i = k_0(x, x_i)$, $K(x)_{i,j} = k(x_i, x_j)$. As we will see, we can use this posterior to build an acquisition function to select the next experiment.

\subsection{Independent Gaussian Processes}
The first model we consider is the simplest. We fit an independent GP to each model. That is, we assume the prior:
\begin{equation}
    f^{(m)} \sim \mathcal{GP}(\mu_0^{(m)}, k_0^{(m)}),
\end{equation}
and obtain the posterior independently, that is, for each fidelity $m = 1, ..., M$, the posterior is obtained by only conditioning on each fidelities data-set: $f^{(m)} | \cup_{m' = 1}^M D^{(m')} = f^{(m)} | D^{(m)}$.
There are two main benefits to this approach: (a) This is the simplest and computationally cheapest out of all exact inference GP methods. (b) It is the easiest on which to intuitively impose prior knowledge and information. However, the main drawback is significant: there is no transfer of information between the fidelities, so we will be forced to make simplifying assumptions relating to hierarchy and bias at the time of optimization.

\subsection{Multi-task Gaussian Processes}

A more complete approach seeks to \textit{jointly} model all fidelities. One effective and popular approach is to use a multi-task Gaussian Process \citep{alvarez2012kernels}. This approach models each fidelity as an output of a Multi-output Gaussian Processes (MOGP). Instead of having a scalar-valued mean function, we have a $M$-dimensional vector valued function, $\mu : \mathcal{X} \rightarrow \mathbb{R}^M$ and a $M \times M$-dimensional matrix valued co-variance function $K : \mathcal{X} \times \mathcal{X}: \rightarrow \mathbb{R}^{M \times M}$, where the $(m, m')$th entry of $K(x, x')$ corresponds to the covariance of $f^{(m)}(x)$ and $f^{(m')}(x')$.

We focus on separable kernels and on the Linear Model of Coregionalization \citep{journel1976mining,goovaerts1997geostatistics}. A \textit{separable kernel} is where the input-dependence and the fidelity dependence can be separated into a product:
\begin{equation}
    K(x, x')_{m, m'} = k^{(I)}(x, x') k^{(F)}(m, m')
\end{equation}
where $k^{(I)}$ and $k^{(F)}$ are scalar kernels corresponding to the inputs and fidelities correspondingly. In matrix notation this can be written as:
\begin{equation}
    K(x, x') = k_I(x, x') \mathcal{B}
\end{equation}
where $k_I$ is now a matrix-valued kernel, and $\mathcal{B}$ is a matrix weighting the task-dependencies. The sum of kernels is also a valid kernel, therefore we can create more flexible models by considering the \textit{sum} of $W$ separable kernels:
\begin{equation}
    K(x, x') = \sum_{w = 1}^W k^{(I)}_w(x, x') \mathcal{B}_w \label{eq: sum_of_separable}
\end{equation}

This allows us to have different types of kernels in the sum, or kernels with different hyper-parameters within the same model. The class of MOGP models with kernels given as (\ref{eq: sum_of_separable}) is known as the Linear Model of Coregionalization (LMC). Special cases of the model include the Intrinsic Coregionalization Model \citep{goovaerts1997geostatistics} (where all kernels $k_w$ are the same) and the Semi-parametric Latent Factor Model \citep{teh2005semiparametric}. 

The main benefit of using these models is that the task-dependency is learnt from the data! We learn to transfer information from the lower-fidelities to the target fidelity. The drawbacks against independent models are (i) higher computational expense when doing exact inference and (ii) the kernel hyper-parameters become difficult to set manually.

\subsection{Further Extensions}

This paper limits itself to the methods mentioned above. However, we mention a few more complicated models that could be used when needed.
The LMC is limited because it assumes that the co-variance function is a linear combination of separable kernels. A more general model allows for non-separable kernels by considering processes convolutions \citep{ver1998constructing, higdon2002space, van2017convolutional}. Deep Gaussian Processes \citep{damianou2013deep} have also been used to model multi-fidelity systems, where each layer represents a fidelity \citep{cutajar2019deep}. Recently, \citet{savage2022deep} used a variant of this method to efficiently run multi-fidelity Bayesian Optimization in a simulated chemical reactor.
\cite{li2021batch} use Bayesian Neural Networks \citep{mackay1992practical, neal2012bayesian} rather than Gaussian Processes, thereby allowing for learning of complex non-linearities while still providing uncertainty estimates. However, such models are notoriously difficult to train \citep{izmailov2021bayesian}. We were interested in applying Bayesian Neural Networks models, but, we were unable to consistently obtain good inference so we discarded them from our analysis.

\section{Bayesian Optimization}

Bayesian Optimization (BO) considers the problem of finding:
\begin{equation}
    x_* = \argmax_{x \in \mathcal{X}} f(x)
\end{equation}
where $f$ is a black box function, and $\mathcal{X}$ is some $d$-dimensional input space. The function can be evaluated at any arbitrary point, $x \in \mathcal{X}$, and this leads to noise corrupted observations of the form:
\begin{equation*}
    y = f(x) + \epsilon
\end{equation*}
where $\epsilon$ is zero-mean noise. BO places a surrogate model on $f$ and uses this model to generate an acquisition function, $\alpha: \mathcal{X} \rightarrow \mathbb{R}$. This function is used to select the sequence of inputs, i.e., to design subsequent experiments. Given a data set $D_t = \{(x_i, y_i) \}_{i = 1}^t$, we select a new point, $x_{t+1}$, by optimizing the acquisition function:
\begin{equation*}
    x_{t+1} = \argmax_{x \in \mathcal{X}} \alpha(x ; D_t)
\end{equation*}

\subsection{Multi-fidelity Bayesian Optimization}

One of the main ideas in BO is managing the trade-off between exploration and exploitation. The algorithm has to decide between experimenting in promising areas where it has observed high values of $f$, against choosing to experiment in areas where there is little information. If there is too much exploration, the algorithm will take too long to find the optimum. However, if there is too much exploitation, the algorithm might settle in a sub-optimal area.

More concretely, at time $t$, we assume that we can \textit{choose} an input-fidelity pair, $(x_t, m)$, from which we obtain a possibly noise observation, $y_t^{(m)}$ of the $m$th fidelity function, $f^{(m)}$, as in Equation (\ref{eq: fidelity_equation}).

And the target of the optimization is to maximize the function at the highest fidelity $f^{(M)}$. We further assume that each fidelity will have a known cost $C^{(m)}$, which is lower than the cost at the highest fidelity. For the purposes of this paper, we will assume the cost refers to a measure of \textit{time}. However, it could represent financial costs, man-power, etc... We clarify that we assume the costs are scalar-values known a priori. While it would be useful to consider a trade-off between different types of costs, this is left to future work.

Multi-fidelity methods allow us to use cheap approximations to explore while saving the expensive evaluations for a separate exploitation phase of the algorithm. All multi-fidelity methods follow the same pattern: at first they will query the lower fidelities and as the optimization loop progresses we begin to query the target fidelity.

\subsubsection{Multi-fidelity Upper Confidence Bound}

\cite{kandasamy2019multi} propose one of the most intuitive and simplest methods. They show that by making the right assumptions, you are able to use independent Gaussian Processes for each fidelity and still transfer information effectively by using Upper Confidence Bounds (UCB). The UCB acquisition function, for a Gaussian Process posterior with mean function, $\mu_t(\cdot)$, and variance function $\sigma_t^2(x)$ is given by:
\begin{equation}
    \alpha_{UCB}(x ; \beta_t, D_t) = \mu_t(x) + \beta_t^{1/2} \ \sigma_t(x)
\end{equation}
In this equation, $\mu_t(x)$ represents our \textit{belief} of how the function looks like. We then add $\beta_t^{1/2}\sigma_t(x)$, which represents the uncertainty in our model, to generate an upper confidence bound of the black-box function. Here, $\beta_t > 0$ is a hyper-parameter that balances how much we value the uncertainty when deciding which point to choose next. A large value represents being optimistic about areas where we are uncertain and will naturally lead to more exploration and reduced exploitation. In practice, $\beta_t$ is usually chosen as an increasing logarithmic function which allows us to have theoretical guarantees. 

Assume that there are $M$ different fidelities, and we are interested in optimizing the highest fidelity, $f^{(M)} := f$. The key assumption is that there is a decreasing \textit{known} maximum bias among the fidelities, such that:
\begin{equation} \label{eq: decreasing bias}
    || f^{(m)} - f \ ||_\infty := \max_{x \in \mathcal{X}} |f^{(m)}(x) - f(x) |\leq \zeta_{MAX}^{(m)}
\end{equation}
For $\zeta_{MAX}^{(1)} \geq \zeta_{MAX}^{(2)} \geq ... \geq \zeta_{MAX}^{(M)} = 0$. That is, we assume that we know the \textit{worst possible} bias that each fidelity might have, and this worst possible bias is strictly non-increasing on fidelity space.

To combine information from multiple fidelities, we begin by fitting an independent GP to the data for each fidelity. That is, for each fidelity $m$, there is a data-set $D_t^{(m)}$ of experiments, so that $f^{(m)} \ | \ D_t^{(m)} \sim \mathcal{GP}(\mu_t^{(m)}, k^{(m)}_t)$. Then the upper confidence bound is defined as:
\begin{equation} \nonumber
    \alpha^{(m)}(x ; D^{(m)}_t) = \mu_t^{(m)}(x) + \beta_t^{1/2} \ \sigma_t^{(m)}(x) + \zeta^{(m)}_{MAX}
\end{equation}
where $(\sigma_t^{(m)})^2(x) = (k^{(m)}_t)^2(x)$. That is, for each fidelity, we sum the mean prediction, the uncertainty, \textit{and the bias}, to obtain an upper bound on the \textit{objective} fidelity. Therefore we can build a narrow bound by considering the minimum of all $M$ upper confidence bounds at each possible input $x$. With this idea in mind, the acquisition function is then defined as:
\begin{equation} 
    \alpha(x ; D_t ) = \min_{m \in \{1, ..., M \}} \alpha^{(m)}(x ; D^{(m)}_t) \label{eq: mf-af}
\end{equation}

\begin{figure*}[ht]
	\centering
	\begin{subfigure}[t]{0.49\textwidth}
	\includegraphics[width = \textwidth]{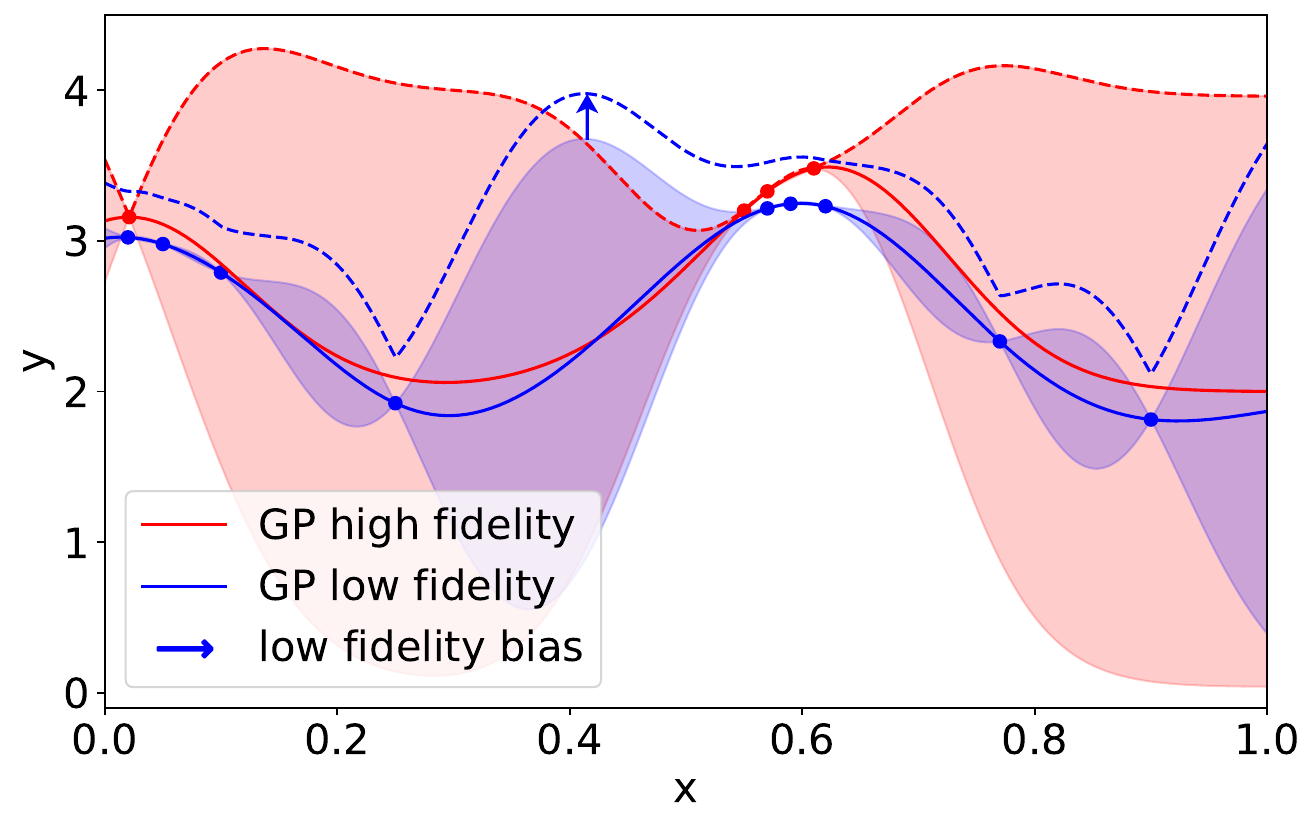}
	\caption{Independent UCBs for each fidelity.}
	\end{subfigure}
	\hfill
	\begin{subfigure}[t]{0.49\textwidth}
	\includegraphics[width = \textwidth]{ 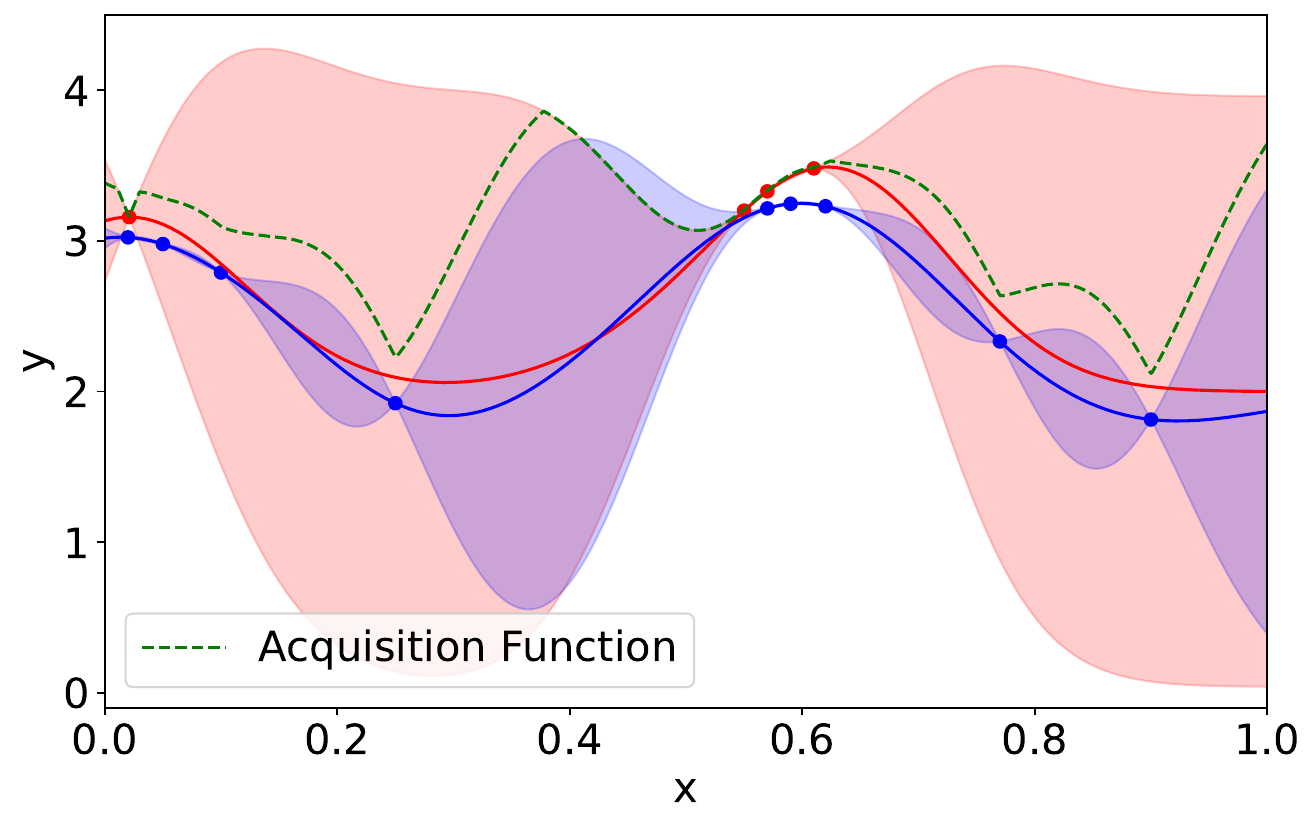}
	\caption{Combination of UCBs to build acquisition function.}
	\end{subfigure}
	\caption{Building a multi-fidelity UCB, in the case where $M = 2$. The dotted lines represent upper confidence bounds based on bias and variances. In the first figure we can see the bounds for each independent fidelity, in the second figure both bounds are combined to obtain the final acquisition function.}
	\label{fig: building mf-af}
\end{figure*}

That is, each $\alpha^{(m)}$ represents the $m$th UCB, and we simply take the minimum to obtain the tightest bound. Figure \ref{fig: building mf-af} gives a graphical example of building this acquisition function.

The next experiment is chosen by maximizing the acquisition function given by (\ref{eq: mf-af}). However, the choice of a fidelity level still remains. Consider the standard deviation of the surrogate model for fidelity $m$ at input $x_{t+1}$, $\sigma_t^{(m)}(x_{t+1})$. If this quantity is high, it means there remains a lot of information to learn about fidelity $m$ at our specific input. If this quantity is low, it means we cannot learn much from experimenting at this fidelity.

Therefore we define a set of threshold levels $\{ \gamma^{(m)} \}_{m=1}^M$. If the posterior variance is smaller than the corresponding threshold, we cannot learn much about the objective function by sampling at this lower fidelity so it is better to choose a higher fidelity. To try to minimize the cost of evaluation, we choose the lowest fidelity that contains enough information according to the thresholds. More precisely, the choice is given by:
\begin{equation} \label{eq: ucb_fidelity_rule}
    m_{t+1} = \argmin_{m \in \{0, ..., M \}} \{ m : \beta^{1/2}_t \sigma_t^{(m)} > \gamma^{(m)} \text{ or } m = M \}
\end{equation}

We note that this idea for choosing the fidelity is not restricted to upper-confidence bound acquisition functions. In fact, we can use \textit{any} acquisition function and then choose the fidelity with rule defined by equation (\ref{eq: ucb_fidelity_rule}). We shall see that combining this rule with LMCs means we can drop the bias assumption and obtain better results than using independent GPs.

\subsubsection{Multi-fidelity Max Entropy Search}

Information-based methods have become increasingly popular in Bayesian Optimization due to their strong theoretical backing. Entropy search \citep{villemonteix2009informational, hennig2012entropy, hernandez2014predictive} tries to maximize the information gain about the optimum input, $x_*$. Unfortunately, this means estimating an expensive $d$-dimensional integral using Monte Carlo approximations. Max-value Entropy Search (MES) \citep{wang2017max} instead focuses on maximizing the information gain about the maximum value of the function, $f_* = f(x_*)$:
\begin{equation}
    \alpha_t(x) = I((x, f(x); f_* | D_t) = H(p(y | D_t, x)) - \mathbb{E}_{f_* | D_t}\left[H(p(y | D_t, x, f_*)) \right]
\end{equation}

The left hand side, $I((x, f(x); f_* | D_t)$ represents the mutual information between the next (possibly) evaluated point, and the maximum value of the function $f_*$. It can be calculated by the difference between the current entropy of the predictive distribution of $y$ at input $x$, and the expected entropy over the distribution of the maximum.

The big advantage of MES is that the expectation is reduced from a $d$-dimensional to 1-dimensional integral. The expectation is estimated using Monte Carlo methods since we are able to sample from the distribution $p(f_* | D_t)$ via Thompson Sampling or an approximation via the Gumbel distribution. More recently, \cite{tu2022joint} and \cite{hvarfner2022joint} jointly consider the joint information gain of both the maximum value, and the optimum input.

A multi-fidelity extension of Max-value Entropy Search (MF-MES) was introduced by \cite{takeno2020multi}. We easily extend to the multi-fidelity setting by considering a joint model, such as the LMC and then maximizing the cost-weighted information gain:
\begin{equation}
    \alpha_t(x, m |D_t) = \frac{I(x, f^{(m)}(x);f_* | D_t)}{C^{(m)}} = \frac{I(x, f^{(m)}(x);f_* | D_t)}{C^{(m)}}
\end{equation}

Here $f_*$ represents the maximum at the \textit{highest} fidelity. So we calculate how much more information we gain about the black-box function's maximum, by querying at input $x$ and fidelity $m$. We divide this by the cost of querying fidelity $m$.

The biggest strength of this method, when compared to multi-fidelity upper confidence bound, is that we are not longer using independent GPs as a model. As such, we can \textit{learn} the relationship between the functions and there is no need to make assumptions about the maximum bias. There are also no hyper-parameters in the acquisition function.

\subsection{Asynchronous Batch Bayesian Optimization}

So far we have focused on sequential Bayesian Optimization. However, in some battery experiments, we cannot wait for the experiment results to return before choosing the next point, we must do this in batch. It is also important to note that we expect different fidelity levels to have a considerably different evaluation time. This can be a problem as most batch methods focus on the synchronous setting. That is, you select a batch, wait for all evaluations to return, and then select another. Large time discrepancies in the evaluation times pose a problem, as there are a lot of resources that will be left idle while longer evaluations finish. As such, we are more interested in the less-studied asynchronous setting.

Throughout this section, assume we are in the normal setting of Bayesian Optimization, however, further assume that at time $t$ we are evaluating a batch of points $\mathcal{Q}_t = \{x_{t, q}\}_{q = 1}^Q$ whose observations, $f_{\mathcal{Q}}$, we still do not know. We are then interested in choosing the next experiment while taking $\mathcal{Q}$ into account, so that the acquisition function depends on the pending evaluations:
\begin{equation}
    x_{t+1} = \argmax_{x \in \mathcal{X}}{\alpha_t(x | D_t, \mathcal{Q}_t)}
\end{equation}

\subsubsection{Thompson Sampling}
Thompson Sampling is a common technique in Bayesian decision making that uses samples from the posterior distribution to guide actions. \citet{pmlr-v84-kandasamy18a} show it can be used effectively for Asynchronous Batch BO. A sample is selected by maximizing a sample from the posterior GP:
\begin{align}
    f_{sample} &\sim \mathcal{GP}(\mu_t, \kappa_t |D_t) \nonumber \\
    x_{t+1} &= \argmax_{x \in \mathcal{X}}{f_{sample}(x)}
\end{align}
Note that the new point selected is \textit{independent} of $\mathcal{Q}$, however, the randomness in the sampling is enough to ensure diversity in the batch. Due to the independence, it works well in the asynchronous batch setting, and large batches can be created cheaply.

\subsubsection{Parallel Querying Through Fantasies}
\cite{snoek2012practical} allow for the asynchronous batching of any acquisition function. This can be achieved by creating multiple `fantasies' at every point in $\mathcal{Q}$, and then using them to marginalize out the pending observations, $f_\mathcal{Q}$:
\begin{align}
    \alpha_t(x | D_t, \mathcal{Q}_t) &= \int \alpha(x | D_t, f_\mathcal{Q}, \mathcal{Q}_t) p(f_{\mathcal{Q}} | D_t) \text{d}f_{\mathcal{Q}} \nonumber \\
    &\approx \frac{1}{S}\sum_{s = 1}^S \alpha_t(x | D_t \cup \{\Tilde{f}^{(s)}_{\mathcal{Q}}, \mathcal{Q}_t \})
\end{align}
where $\Tilde{f}^{(s)}_{\mathcal{Q}}$ are sample drawn from the normal distribution $f_\mathcal{Q} | D_t$, and $S$ is the number of samples. Note that we are trying to estimate a $Q$-dimensional integral, so the larger the number of fantasies, the better the approximation but the computational expense increases. One consideration though, is that we \textit{do not} need to fully re-train the Gaussian Process for every fantasy. This is because the variance of the posterior only depends on the input locations, $\mathcal{Q}$, which are the same for every fantasy. Instead, we only need to re-train the mean of the process which is much cheaper.

In the case of Max-value Entropy Search, \cite{takeno2020multi} show that using fantasies, the $Q$-dimensional integral can be written as a 2-dimensional integral which means we do not need as many fantasies to get good approximations. 

Gaussian Process Bandit UCB \citep{desautels2014parallelizing} provides a computationally cheap version of fantasizing, where a single sample, equal to the predictive mean of the GP at all points in $\mathcal{Q}$ is used, i.e., you assume that all pending observations will simply equal the predictive mean (and so the predictive mean remains unchanged, while the predictive variance decreases around pending points). Recently, \citet{zhang2022machine} showed its practicality by using to design bacterial ribosome binding sites. We do note that the empirical success of this method is restricted to the Upper Confidence Bound BO, and may not work as well with other acquisition functions.

\subsubsection{Local Penalization}

Local penalization is a batch Bayesian Optimization algorithm introduced by \cite{gonzalez2016batch}, and which was then extended to the asynchronous setting by \cite{alvi2019localpen}. 

\begin{figure}
    \centering
    \includegraphics[width = 4in]{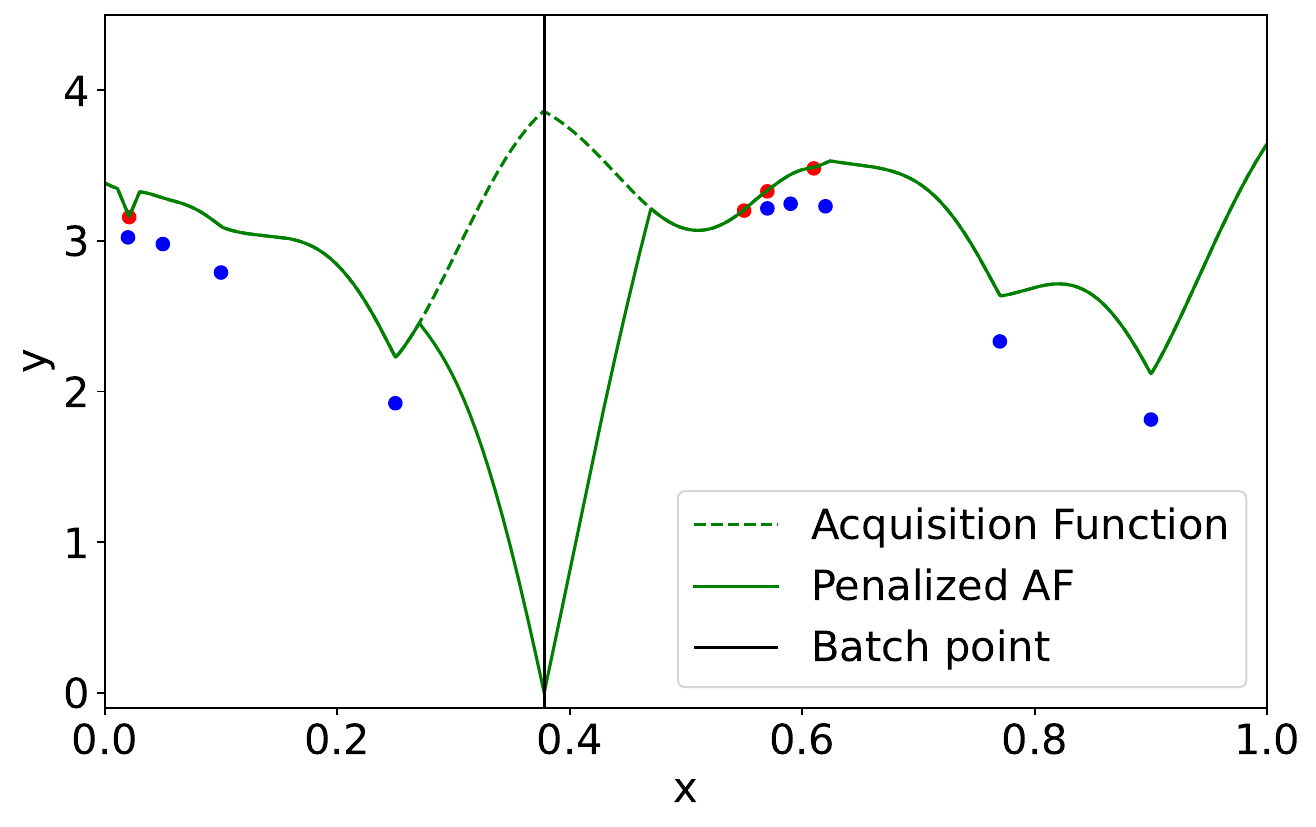}
    \caption{An example of the local penalization process. We selected $x = 0.378$ as a point to evaluate. The acquisition function is then penalized as shown above until we obtain a new observation. This allows us to select new queries without receiving an observation, as the penalized acquisition function is guaranteed to have a different maximum. The blue and red dots represent low and high fidelity observations, respectively.}
    \label{fig: batch_demo}
\end{figure}

The method relies on trying to mimic the sequential setting, where after receiving experimental information about a particular input we usually expect the acquisition function to decrease in the neighbourhood surrounding this point. More precisely, assume we wish to select a new point $x_{t}$, then we can mimic the sequential setting by defining the new acquisition function, $\Tilde{\alpha}$, as:
\begin{equation}
    \Tilde{\alpha}(x ; D_t, \mathcal{Q}) = g(\alpha(x ; D_t)) \prod_{x_{t, q} \in \mathcal{Q}} \psi( x_{t, q} ) \label{eq: local_pen}
\end{equation}
where $\psi(x)$ is a function that penalizes the region around $x$, and $g(\cdot)$ is an increasing differentiable transformation that makes sure the acquisition function is positive. And so, in Equation \ref{eq: local_pen} we have a positive transformation of the original acquisition function, multiplied by penalizers centered at every pending experiment. We use $g(z) = z$ for positive acquisition functions, and the soft-max transformation otherwise, $g(z) = \log(1 + e^{z})$. The question then becomes, how do we choose such penalizers? To do this, we begin by assuming that the function satisfies the Lipschitz condition. That is, if $\mathcal{X}$ is compact, then:
\begin{equation*}
    | f(x_1) - f(x_2) | \leq L || x_1 - x_2 || \quad \forall x_1, x_2 \in \mathcal{X}
\end{equation*}
where $L \in \mathbb{R}^+$, and $||\cdot||$ is the $L^2$ norm in $\mathbb{R}^d$. We now consider the ball, around arbitrary $x_j \in \mathcal{X}$:
\begin{equation*}
    B_{r_j}(x_j) = \{x \in \mathcal{X} : ||x_j - x || \leq r_j \} 
\end{equation*}
where the radius $r_j$ is given by:
\begin{equation*}
    r_j = \frac{f(x_*) - f(x_j)}{L} =: r(x_j)
\end{equation*}
The radius is chosen because, if the Lipschitz assumption holds, then $x_*$ cannot lie inside this ball. Note that due to $f$ being modeled by a GP, then this radius is a random quantity. \cite{gonzalez2016batch} then define the local penaliser as follows:
\begin{equation}
    \psi(x; x_j) = 1 - p(x \in B_{r_j}(x_j)) \nonumber
\end{equation}
That is, we choose the local penalizer to be the probability that $x$ does not lie inside $B_{r_j}(x_j)$. However, \cite{alvi2019localpen} argue that this penalizer may lead to redundantly sampling the same points. Instead they propose the penalizer:
\begin{equation}
	\psi(x; x_j) = \min \left\{ \frac{|| x - x_j ||}{\mathbb{E}(r_j) + \sigma_t(x_j) / L}, \ 1 \right\} \label{eq: br_penaliser}
\end{equation}
Figure \ref{fig: batch_demo} shows an example of how the penalization can be used to alter the acquisition function to choose a new experiment.

\subsection{Relationship between multi-fidelity and asynchronous batch methods}

In Table \ref{tab: fid_and_batch_table} we can find a summary of all the introduced methods for multi-fidelity and batching, including information about the weaknesses and strengths of each method. 

Ultimately the goal of using multi-fidelity and batch methods is to speed up the optimization by using extra tools available to us. Multi-fidelity methods leverage cheap and approximations to the objective, while batching takes advantage of running experiments in parallel. Therefore it is natural to want to use both methods at the same time to obtain even more speed gains in optimization procedures.

However, we emphasise that we cannot fully use both tools simultaneously without asynchronicity arising. Indeed, by definition, lower fidelity approximations are meant to be cheaper and faster to obtain. This means if we select a batch of evaluations, those of lower fidelities will finish earlier than high fidelity evaluations. Without any asynchronicity we must wait for all experiments to finish before deciding the next batch, making the method inefficient.

There can also be strong relationships between fidelity and \textit{batch size}. Indeed, it might be the case that many low fidelity experiments can easily be done in parallel, but due to having to use more resources, for high fidelities we can only parallelize a few. An example of this could be as simple as trying out cake recipes. A full fidelity observation would be baking a cake in the oven; while low fidelity observations could be making muffins instead of a full cake. In such an example, since we are limited by oven size, we can parallelize low-fidelity experiments much easier than high fidelity ones.

We focus on multi-fidelity and batch methods that allow mimicking of sequentially choosing experiments and fidelities, this means we can select which experiments to run without worrying about selecting a batch size or limiting ourselves to single-fidelity batches.

\begin{table}[h!]
\centering
\begin{tabular}{|p{2.5cm}|p{3.5cm}|p{4.3cm}|p{4.3cm}|}
\hline
\multicolumn{4}{|c|}{\textbf{Multi-fidelity}} \\
\hline
\textbf{Method} & \textbf{Reference} & \textbf{Strengths} & \textbf{Weaknesses} \\ [0.5ex]
\hline
\begin{tabular}[t]{@{}c@{}} MF-GP-UCB \end{tabular}
& \begin{tabular}[t]{@{}c@{}}\cite{kandasamy2019multi} \end{tabular} 
& \begin{tabular}[t]{@{}l@{}} 
(a) The acquisition function \\
and fidelity criterion can be  \\
calculated cheaply and exactly \\
(for data-sets of a few hundred \\
experiments it should take a \\
few seconds to optimize) \smallskip \\
(b) Using independent GPs \\
allows for expert knowledge \\
to be incorporated more \\
easily. \smallskip \end{tabular} 
& \begin{tabular}[t]{@{}l@{}}
(a) It requires the maximum \\
bias to be known. By using \\
independent GPs it does not \\
learn relationships between \\
fidelities. \smallskip \\
(b) We need to choose the \\
value of the threshold \\ 
hyper-parameter and for the \\
UCB acquisition function. \end{tabular} \\
\hline
\begin{tabular}[t]{@{}c@{}} MF-MES \end{tabular}
& \begin{tabular}[t]{@{}c@{}} \cite{takeno2020multi} \end{tabular} 
& \begin{tabular}[t]{@{}l@{}}
(a) Relationship between \\ 
fidelities is learnt from data. \\
Allowing for more efficient \\ 
experimental choices. \smallskip \\
(b) No hyper-parameters \\ 
to choose outside of those \\
used in approximation \\
techniques. \smallskip \\
(c) Batch extension only \\
requires calculation of \\
2-dimensional integral. \end{tabular}  
& \begin{tabular}[t]{@{}l@{}}
(a) The acquisition function \\
and fidelity criterion cannot be  \\
calculated exactly and needs to \\
be estimated. Computational \\ 
cost may grow depending on \\ 
the approximation quality. \\
(it could take up to a few \\
minutes to optimize) \smallskip \\
(b) Suffers from normal BO \\
drawbacks: e.g. over- \\ 
exploration of boarders \smallskip \\
(c) Cannot incorporate expert \\ 
knowledge easily. \smallskip \end{tabular} \\
\hline
\hline
\multicolumn{4}{|c|}{\textbf{Asynchronous Batch}} \\
\hline
\textbf{Method} & \textbf{Reference} & \textbf{Strengths} & \textbf{Weaknesses} \\ [0.5ex]
\hline
\begin{tabular}[t]{@{}c@{}} Thompson \\
Sampling \end{tabular} 
& \begin{tabular}[t]{@{}c@{}}\cite{pmlr-v84-kandasamy18a} \end{tabular} 
& \begin{tabular}[t]{@{}l@{}}
(a) Parallel computation of \\
batch points allows for cheap \\
creation of large batches (but \\
computational cost can be \\
very variable depending on \\
coarseness of sampling grid) \smallskip \\
(b) Hyper-parameter free \smallskip \end{tabular}  
& \begin{tabular}[t]{@{}l@{}}
(a) Experiment designs can \\
be too exploitative (i.e. \\
redundant in batch) \smallskip \\
(b) Calculation can be \\
unfeasible for grids with \\
a large number of points
\end{tabular} \\
\hline
\begin{tabular}[t]{@{}c@{}} Fantasies \end{tabular} 
& \begin{tabular}[t]{@{}c@{}}\cite{snoek2012practical} \end{tabular} 
& \begin{tabular}[t]{@{}l@{}}
(a) Strong theoretical \\ 
motivation \smallskip \\
(b) No hyper-parameters \\ 
to choose outside of those \\
used in approximation \\
techniques. \smallskip \end{tabular}  
& \begin{tabular}[t]{@{}l@{}}
(a) Acquisition function is \\
a large dimensional integral. \\
Approximation may be\\
unfeasible for large batch \\ 
sizes. \smallskip \\
(b) As such, acquisition \\
function may take a long time \\
to optimize (from a few \\
minutes to a few hours) \smallskip \end{tabular} \\
\hline
\begin{tabular}[t]{@{}c@{}} Local \\
Penalization \end{tabular} 
& \begin{tabular}[t]{@{}c@{}}\cite{gonzalez2016batch} \\
\cite{alvi2019localpen} \end{tabular} 
& \begin{tabular}[t]{@{}l@{}}
(a) Acquisition function \\
can be calculated analytically \smallskip \\
(b) Heuristic is well motivated, \\
and has strong empirical \\
performance. \end{tabular}  
& \begin{tabular}[t]{@{}l@{}}
(a) Requires approximation \\
of various hyper-parameters\smallskip \\
(b) Batch creation is done \\
sequentially, so it can \\
be expensive for large \\
batch sizes (a few minutes \\
to optimize) \end{tabular} \\
\hline
\end{tabular}
\caption{Comparison table for multi-fidelity and asynchronous batch methods. Highlighting and summarizing the strengths and weaknesses of each method.}
\label{tab: fid_and_batch_table}
\end{table}

\section{Asynchronous Multi-fidelity Batch Optimization}

\subsection{Problem Setting}

We now seek to formalize the problem setting. The method presented seeks to find the best possible solution, as \textit{quickly} as possible. We are interested in:
\begin{equation*}
    x_* = \argmax_{x \in \mathcal{X}} f(x)
\end{equation*}
Where $\mathcal{X}$ denotes a \textit{continuous} input space. We assume $f$ is expensive to evaluate, but, we have access to lower fidelity approximations, $f^{(m)}$, which tend to be cheaper (for $m = 1, ..., M$). These approximations may be biased or noisy.

Because we are trying to optimize with respect to time, we introduce a discrete time component. We denote the input of the $i$th experiment by $x_{i, t}^{(m)}$, where $t \in \{1, ..., T \}$ denotes the time at which the experiment began, and $m \in \{1, ..., M \}$ denotes the fidelity.

When we select an input point, $x_{i,t}^{(m)}$, we obtain a noisy observation of $f^{(m)}$. However, we may not observe it at time $t$. Instead, we observe it at time $t + \tau_i^{(m)}$, where $\tau_i^{(m)}$ can be a deterministic or random time. We denote the corresponding observation $y_{i, t + \tau_i}^{(m)}$. We assume the relationship between the observation and the experimental input is given by:
\begin{equation}
	y_{i, t + \tau_i}^{(m)} = f^{(m)}\left(x_{i,t}^{(m)}\right) + \eta^{(m)} \epsilon_i,
\end{equation}
where $\eta^{(m)}$ denotes the noise level. We assume $\epsilon_i$ is standard Gaussian, where the variance may depend on the fidelity level. 

Finally we assume that each fidelity has an associated batch space $\lambda^{(m)}$ (which in practice typically increases with fidelity). We assume this batch space is not not freed up until we have an observation. We also have a maximum budget which we cannot exceed at any one time, $\Lambda$. If we let $\lambda_i$ denote the batch space associated with the $i$th observation, and we let $\mathcal{I}$ denote the index set of the unobserved queries, we can write this as the constraint:
\begin{equation}
	\text{TotalBatchSpace}_t := \sum_{i \in \mathcal{I}} \lambda_i \leq \Lambda
\end{equation}
This idea behind this constraint is that we may be able to run a lot of low-fidelity observations or a few high-fidelity observations or a mix of both. We note that this batch space is independent of the time it takes to query each fidelity. We are introducing this as to allow variable batch sizes. An example of this in practice, because coin cell batteries are smaller and use less material, we can concurrently run more experiments than for pouch cell experiments.

The \textbf{aim} of the algorithm would be to obtain the best possible approximation of the optimum point at time $T$.

\subsection{A general algorithm for any acquisition function}

A general algorithm can be constructed from all the tools mentioned above. For this we require: (a) A Multi-task Gaussian Processes model, because we will require information transfer between the fidelities, (b) A method that allows for asynchronous batching, (c) A way of choosing which fidelity to query, and (d) An acquisition function. We propose iteratively choosing the points, we first optimize a batch-modified acquisition function \textit{on the target fidelity}:
\begin{equation}
    x_{i+1,\ t} = \argmax_{x \in \mathcal{X}}{\Tilde{\alpha}_t(x, M | D_t, \mathcal{Q}_t)}
\end{equation}
Where $\mathcal{Q}_t$ is the batch of points being evaluated at time $t$, and $\Tilde{\alpha}$ is a acquisition function modified to consider that we are currently evaluating $\mathcal{Q}_t$ by either local penalizing or fantasizing. Once we have chosen the point, we can choose the fidelity. We consider two alternatives, the first one is the heuristic used on MF-GP-UCB:
\begin{equation}
    m_{i+1,\ t} = \argmin_{m \in \{0, ..., M \}} \{ m : \beta^{1/2}_t \sigma_t^{(m)}(x_{i+1, \ t}) > \gamma^{(m)} \text{ or } m = M \} \label{eq: ucb_choose_fidelity}
\end{equation}
Recall, $\beta_t^{1/2}\sigma_t^{(m)}(x_{i+1, \ t})$ measures the posterior uncertainty at fidelity $m$ and input $x_{i+1, \ t}$, and therefore, we choose the smallest fidelity whose uncertainty is larger than $\gamma^{(m)}$.
In other words, we query at the lower fidelities until the uncertainty at each fidelity is low. The second alternative is to consider the information-based criterion introduced by MF-MES:
\begin{equation}
    m_{i+1} = \argmax_{m \in \{0, ..., M \}} \frac{\mathbb{E}_{f_{\mathcal{Q}} | D_t} \left[ I(x, f^{(m)}(x_{i+1, \ t});f_* | D_t, f_\mathcal{Q}, \mathcal{Q}_t) \right]}{\mathbb{E}\left[\tau^{(m)}\right]} \label{eq: mf_mes_choose_fidelity}
\end{equation}
That is, we use the expected information gain of querying the experiment at the $m$th fidelity, where the expectation is taken over the pending observations (and we also weight by the expected cost). The main advantage of pre-selecting the experimental input using another acquisition function, instead of optimizing MF-MES directly, is that we do not need to rely on approximating an information-based acquisition function during the optimization, and instead we only evaluate the expensive acquisition function $M$-times. In addition, it allows us to use specialist acquisition functions, as we shall see in the experiment section. 
As any acquisition function can be selected (including multi-objective acquisition functions), future work can extend the proposed methodology to cases with multiple objectives, which arise in many engineering applications \citep{park2018multi, schweidtmann2018machine, thebelt2022multi,badejo2022integrating, tu2022joint}. 

We then repeat this procedure until the maximum budget is used up. Then, at each time point, we check if there are any new observations. If there are, we update the model, and carry out the previous procedure until the budget is full again. The full procedure is detailed in Algorithm \ref{algo: live_batch}.

\begin{algorithm}
\caption{Asynchronous Multi-fidelity Batch Optimization} \label{algo: live_batch}
\begin{algorithmic}
 \STATE INPUT: Maximum Budget: $\Lambda$. Batch Spaces: $\lambda^{(m)}; \ m = 1, ..., M$. Acquisition Function: $\alpha$. Choice of Asynchronous Bayesian Optimization: Thompson Sampling, Fantasies, or Local Penalization. Choice of fidelity-selector: UCB or Information-based. [Optional] Fidelity Thresholds Parameters: $\gamma^{(m)}; \ m = 1, ..., M$ and $\beta_t$. Number of Fantasies: $S$.
 \STATE Begin algorithm:\\
 \FOR{$t = 1, 2, 3, ..., T$}
 \STATE Obtain observations $\mathcal{Y}_t$ from queries $\mathcal{X}_t$, with index set $\mathcal{I}_t$
 \STATE	$\mathcal{Q}_t = \mathcal{Q}_{t-1} \backslash \mathcal{X}_t$ \texttt{ \# remove observations from batch}
 \STATE	$D_t = D_{t-1} \cup (\mathcal{X}_t \times \mathcal{Y}_t)$ \texttt{ \# expand data set}
 \STATE $\text{TotalBatchSpace}_t = \text{TotalBatchSpace}_{t-1} - \sum_{j \in \mathcal{I}_t}\lambda^{m_j}$ \texttt{ \# free up cost of evaluations}
 \STATE Update Surrogate Model
 \WHILE{$\text{TotalBatchSpace}_t < \Lambda$}
    \IF{BBO Method is Local Penalization}
    \STATE Estimate Maximum Value, $P$, and Lipschitz Constant, $L$
    \STATE Build Initial Acquisition Function $\alpha_t(x, M | D_t)$
    \STATE $\Tilde{\alpha}_t(x, M | D_t) = \alpha_t(x, M | D_t) \prod_{x_q \in \mathcal{Q}}\psi(x_q; \hat{L}, \hat{P})$
    \ELSIF{BBO Method is Fantasies}
    \STATE $\mathcal{S} \leftarrow $ Sample $f_\mathcal{Q} | D_t$ ($S$ times)
    \STATE $\Tilde{\alpha}_t(x, M | D_t) = \sum_{i = 1}^S \alpha_t(x, M | D_t, \mathcal{S}_i, \mathcal{Q}$)
    \ELSIF{BBO Method is Thompson Sampling}
    \STATE  $ f_{sample} \sim \mathcal{GP}(\mu_t, \kappa_t |D_t) $
    \STATE $ x_{t+1} = \argmax_{x \in \mathcal{X}}{f_{sample}(x)} $
    \ENDIF
    \STATE $x_{i, t} = \argmax{\alpha_t(x, M | D_t)}$
    \STATE $m_{i, t} = $ FidelitySelector($x_{i, t+1}$, $D_t$, $\mathcal{Q}$)
    \STATE $\mathcal{Q}_t = \mathcal{Q}_t \cup \{x_{i, t}\}$
    \STATE $\text{TotalBatchSpace}_t = \text{TotalBatchSpace}_t + \lambda^{(m_{i, t})}$
    \STATE $i = i + 1$
\ENDWHILE
\ENDFOR
\end{algorithmic}
\end{algorithm}

\subsection{Parameter Estimation}

In this section we provide brief and heuristic guidance for the estimation of parameters. In particular, all Gaussian Process models have learnable hyper-parameters. In addition, if we use equation (\ref{eq: ucb_choose_fidelity}) for fidelity-selection, then we introduce hyper-parameters relating to the fidelity thresholds, $\beta_t$ and $\gamma^{(m)}$. For Local Penalization, we are also required to estimate the Lipschitz constant and the maximum value of the function. For Fantasizing we need to choose the number of fantasies we want. MES-based fidelity selection are parameter-free, but more computationally costly. For ease of notation, we write $x_i$ to refer to $x_{i, t}^{(m)}$.

\subsubsection{The kernel and noise parameters}

We estimate the parameters of the GP, $\theta$, by maximizing the marginal log likelihood \citep{rasmussen2005gps}:
\begin{equation}
    \log p(y | x, \theta) = - \frac{1}{2} \log | K_{\theta} | - \frac{1}{2}(y - \mu)^T K_\theta^{-1} (y - \mu) - \frac{N}{2} \log (2 \pi) 
\end{equation}

Where $K_\theta$ is the covariance matrix of the training set, and $\mu$ is the predictive mean on the training set. However, we must note that only the lowest fidelity will have observations across the whole search space. The higher fidelities will only get observations in promising areas, meaning that estimating the hyper-parameters directly may lead to inaccuracies when using independent GPs. To counter this, we only fully train the GP hyper-parameters on the lowest fidelity for any multi-fidelity methods. For the other fidelities, we fix the prior mean constant and the length-scales, and we only train the scaling constants and the likelihood noise. We note that the above heuristic is unnecessary for multi-task GPs, where we can train all the kernel hyper-parameters in the combined data-set of all observations

\subsubsection{Choosing the fidelity thresholds, $\gamma^{(m)}$}

Recall that we use the fidelity thresholds to decide in which fidelity to experiment at, and therefore it is very important that they are tuned correctly. If they are too high, we will not use the lower fidelities effectively, but if they are too low we will be stuck in the lower fidelities for a long time. To counter this, \cite{kandasamy2019multi} proposes the following heuristic: initialise the thresholds as small values, and if the process does not experiment above the $m$th fidelity for more than $\tau^{(m+1)} / \tau^{(m)}$ time points, double the value of $\gamma^{(m)}$. If the time intervals are random, we can use expectations instead.

Alternatively, we found good experimental results were achieved by setting $\gamma = 0.1$. However, we note that all benchmarks were quasi-normalized beforehand so that the output values would be close to $1$. In practice fixed thresholds should be carefully scaled with the outputs.

\subsubsection{The penalization parameters}

The local penalisers depend on two parameters, the Lipschitz constant, $L$, and the maximum value of the function, $P = f(x_*)$. In practice, we will never know them, so it is important to have a good method of estimating them. \cite{gonzalez2016batch} propose ways of estimating both.

The maximum value can be estimated cheaply with $\hat{P} = \max_{i}\{y_i\}$, or with the slightly less rough estimate $\hat{P} = \max_{x \in \mathcal{X}} \mu_t(x)$. We tend to prefer the first option, as it is slightly cheaper, and we expect the estimates to be very similar. To adapt it to the multi-fidelity case, we get an estimate independently for each fidelity, keeping in mind we are trying to mimic the sequential case.

Estimating the Lipschitz constant is slightly more elaborate. It can be shown that $L_\nabla = \max_{x \in \mathcal{X}} || \nabla f(x)||$ is a valid Lipschitz constant. In addition, we know the distribution of the gradient:
\begin{equation*}
    \nabla f(x) | D_t \sim \Normal(\mu_{\nabla}(x), \Sigma^2_{\nabla}(x)) 
\end{equation*}
where the mean vector is:
\begin{equation*}
    \mu_{\nabla}(x) = \partial K_{t,*}(x)\Tilde{K}^{-1}_t y
\end{equation*}
For $\Tilde{K}_t = K_t + \eta^2 I$ and $(\partial K_{t,*})_{i, l} = \partial k_t(x) / \partial x^{(i)}$, and $y$ is the vector containing all observations so far. We can then estimate the Lipschitz constant as $\hat{L} = \max_{x \in \mathcal{X}} || \mu_{\nabla}(x)||$.

This could further be improved, \cite{alvi2019localpen} propose using the local estimator, $L_{\nabla, i} = \max_{x \in \mathcal{X}_i} || \mu_{\nabla}(x) ||$, where $X_i$ is a local region around the input of the $i$th experiment. This allows the penalization to adapt to the smoothness of the area it is being evaluated in.

\section{Experimental Results} \label{sec: experiments}

In this section we present our empirical results. As is common in the Bayesian Optimization literature, we use log-regret as a measure of performance, which is defined as:
\begin{equation*}
    \text{log-regret} = \log\left[f(x_*) - \max_{i=1, ..., N_t}f(x_i) \right]
\end{equation*}
where $x_i$ represents the $i$th experiment, and $N_t$ is the number of experiments finished at time $t$. It measures how close our best observation is to the optimum of the function. We take the logarithm to make comparisons easier, since all benchmarks were quasi-normalised so that the outputs are small.

\begin{table}
\noindent
\begin{tabular}{|c||*{4}{c|}}\hline
Method
&\makebox[10em]{Reference}&\makebox[6em]{Multi-fidelity}&\makebox[6em]{Batching}&\makebox[4em]{Model}\\\hline\hline
UCB & \cite{srinivas2009gaussian} & \xmark & \xmark & Single GP\\\hline
MF-GP-UCB & \cite{kandasamy2019multi} & Variance-based & \xmark & Independent GPs \\\hline
MF-GP-UCB w LP & This work & Variance-based & Local Penalization & Independent GPs \\\hline
PLAyBOOK (UCB) & \cite{alvi2019localpen} & \xmark & Local Penalization & Single GP\\\hline
UCB-V-LP & This work & Variance-based & Local Penalization & MOGP\\\hline
UCB-I-LP & This work & Information-based & Local Penalization & MOGP\\\hline \hline
TuRBO-TS & \cite{eriksson2019scalable} & \xmark & Thompson Sampling & Single GP\\\hline
TuRBO-V-TS & This work & Variance-based & Thompson Sampling & MOGP\\\hline
TuRBO-I-TS & This work & Information-based & Thompson Sampling & MOGP\\\hline\hline
MF-MES & \cite{takeno2020multi} & Information-based & Fantasies & MOGP\\\hline
\end{tabular}
\caption{Summary of all the methods compared in this section. We split the table into three sections: the first one refers to UCB-based methods, the second to the our modifications of TuRBO, and for the third we include multi-fidelity Max Entropy Search (MF-MES) as a general baseline. Variance-based and Information-based multi-fidelity refers to using Equations (\ref{eq: ucb_choose_fidelity}) and (\ref{eq: mf_mes_choose_fidelity}) (respectively) to choose which fidelity to query. Local penalization is implemented as described in \citep{alvi2019localpen}, using hard local penalizers and a local estimate of the Lipschitz constant. Thompson Sampling is carried out across a fixed grid in TuRBO trust regions. In all cases, for TuRBO, we use a single trust region. For all non-batching methods, we fill the batch through random sampling as to avoid idle resources. We used GPyTorch and BoTorch to implement the methods \citep{paszke2019pytorch, gardner2018gpytorch, balandat2020botorch}.}
\label{tab: methods_summary}
\end{table}

\subsection{Synthetic Benchmarks}

We focus on implementing the same benchmarks as \cite{kandasamy2019multi} (some of which are taken from \cite{xiong2013sequential}). More details about how each fidelity is created can be found in Section \ref{sec: benchmark_details}. We compare against a simple single-query, single-fidelity UCB \citep{srinivas2009gaussian}, against single-query multi-fidelity Gaussian Processes Upper Confidence Bound (MF-GP-UCB) \citep{kandasamy2019multi}. We also compare against a single-fidelity PLAyBOOK \citep{alvi2019localpen}, which is simply the UCB acquisition function, in conjunction with the hard local penalizers introduced in Equation (\ref{eq: br_penaliser}), and with local estimation of the Lipschitz constant. Finally, we also include results for TuRBO \citep{eriksson2019scalable}, whose details are explained in Section \ref{sec: exp_high_dim}. For standard BO methods we use UCB as our acquisition function, for TuRBO methods we use Thompson Sampling, and we use MF-MES to represent information-based methods. A summary of all benchmarks and their properties can be found in Table \ref{tab: methods_summary}.

\begin{figure}
	\begin{subfigure}{0.49\textwidth}
	\includegraphics[width = \textwidth]{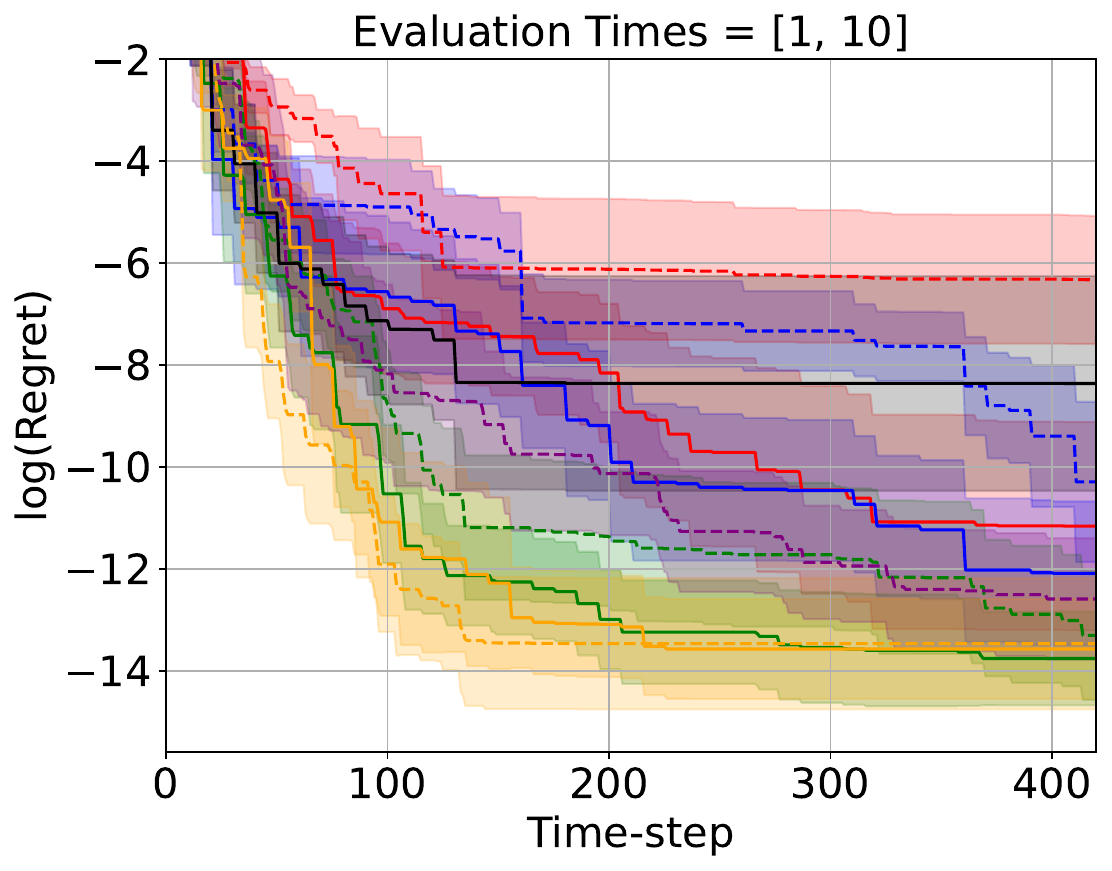}
	\caption{Currin2D}
	\end{subfigure}
	\hfill
	\begin{subfigure}{0.49\textwidth}
	\includegraphics[width = \textwidth]{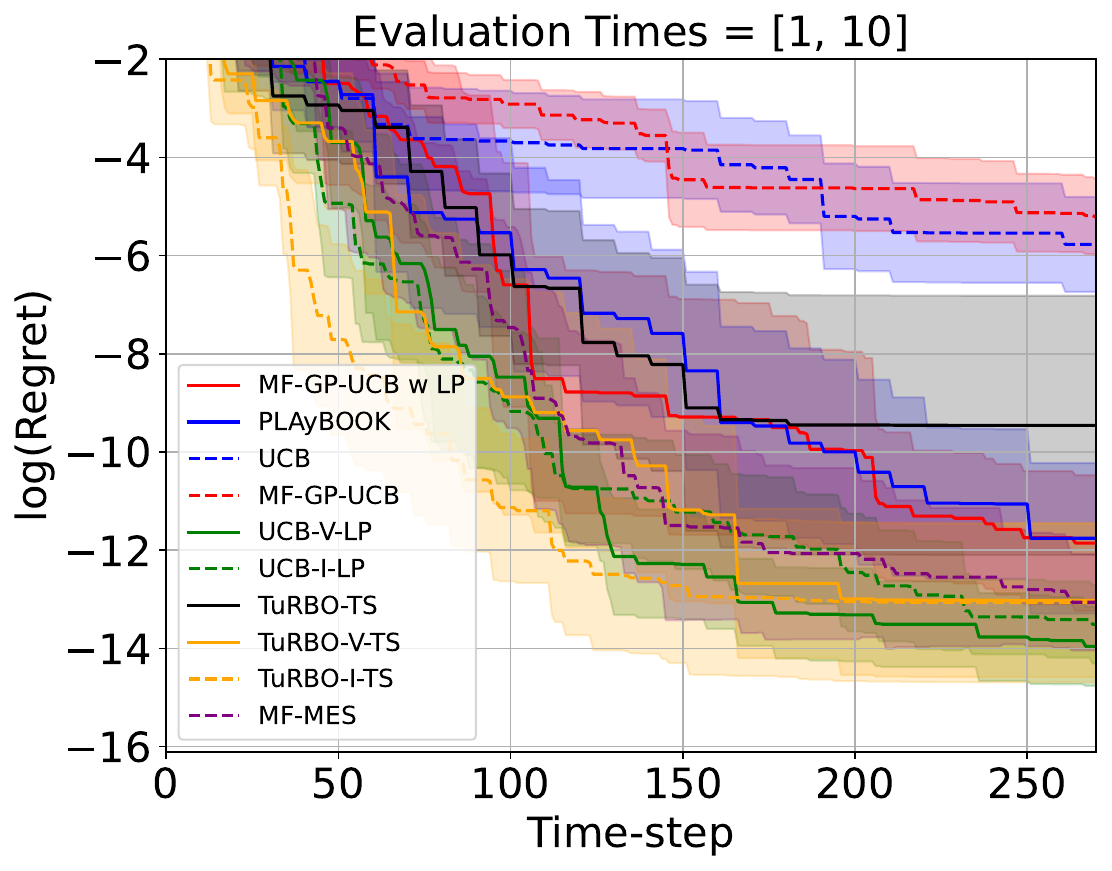}
	\caption{BadCurrin2D}
	\end{subfigure}
	\hfill
	\begin{subfigure}{0.49\textwidth}
	\includegraphics[width = \textwidth]{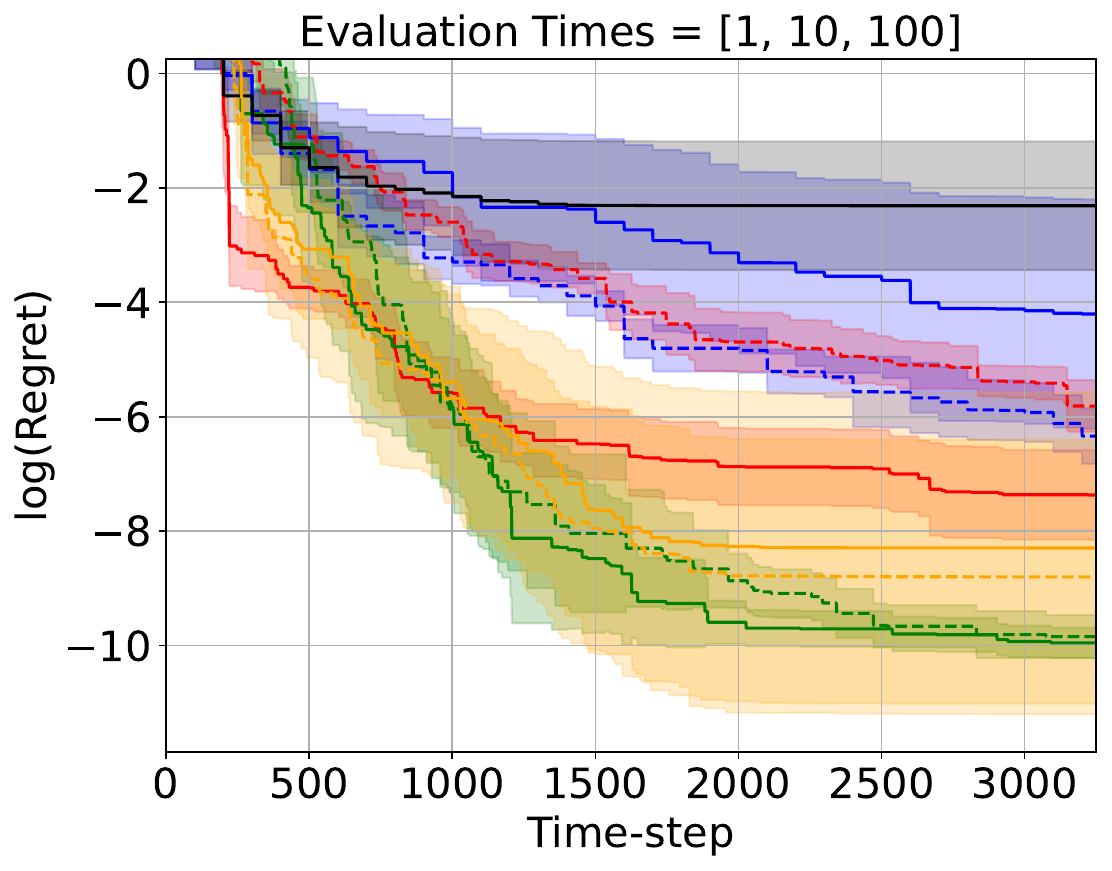}	
	\caption{Hartmann3D}
	\end{subfigure}
	\hfill
	\begin{subfigure}{0.49\textwidth}
	\includegraphics[width = \textwidth]{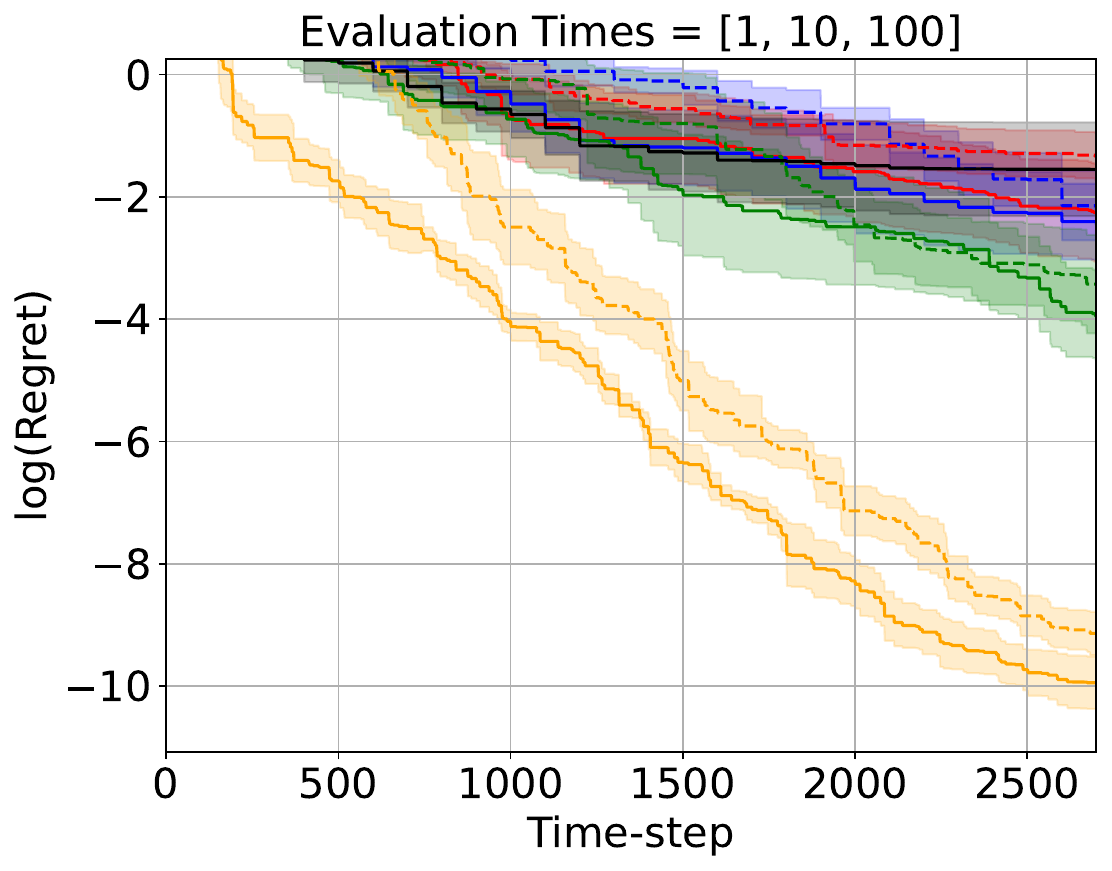}	
	\caption{Hartmann6D}
	\label{fig: sub_hartmann6d}
	\end{subfigure}
	\hfil
	\begin{subfigure}{0.49\textwidth}
	\includegraphics[width = \textwidth]{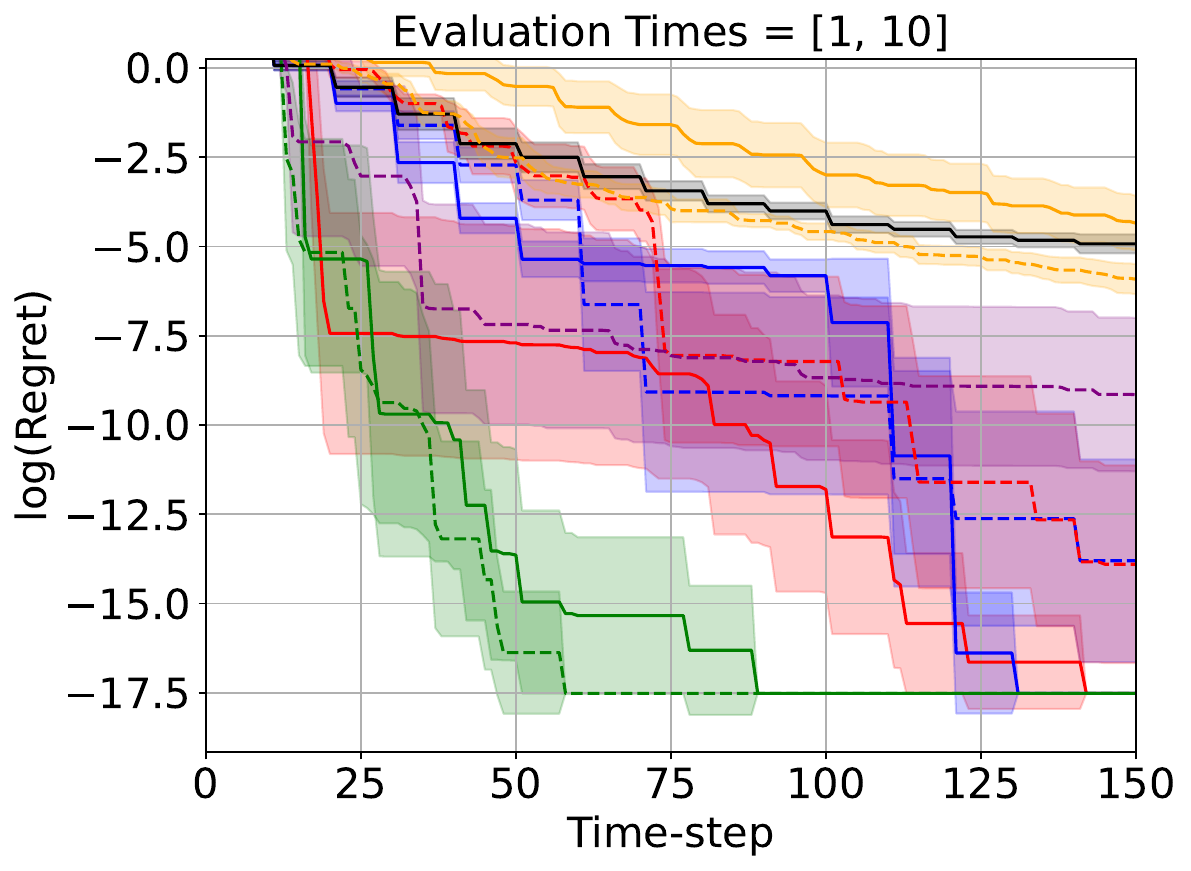}	
	\caption{Park4D}
	\end{subfigure}
	\hfill
	\begin{subfigure}{0.49\textwidth}
	\includegraphics[width = \textwidth]{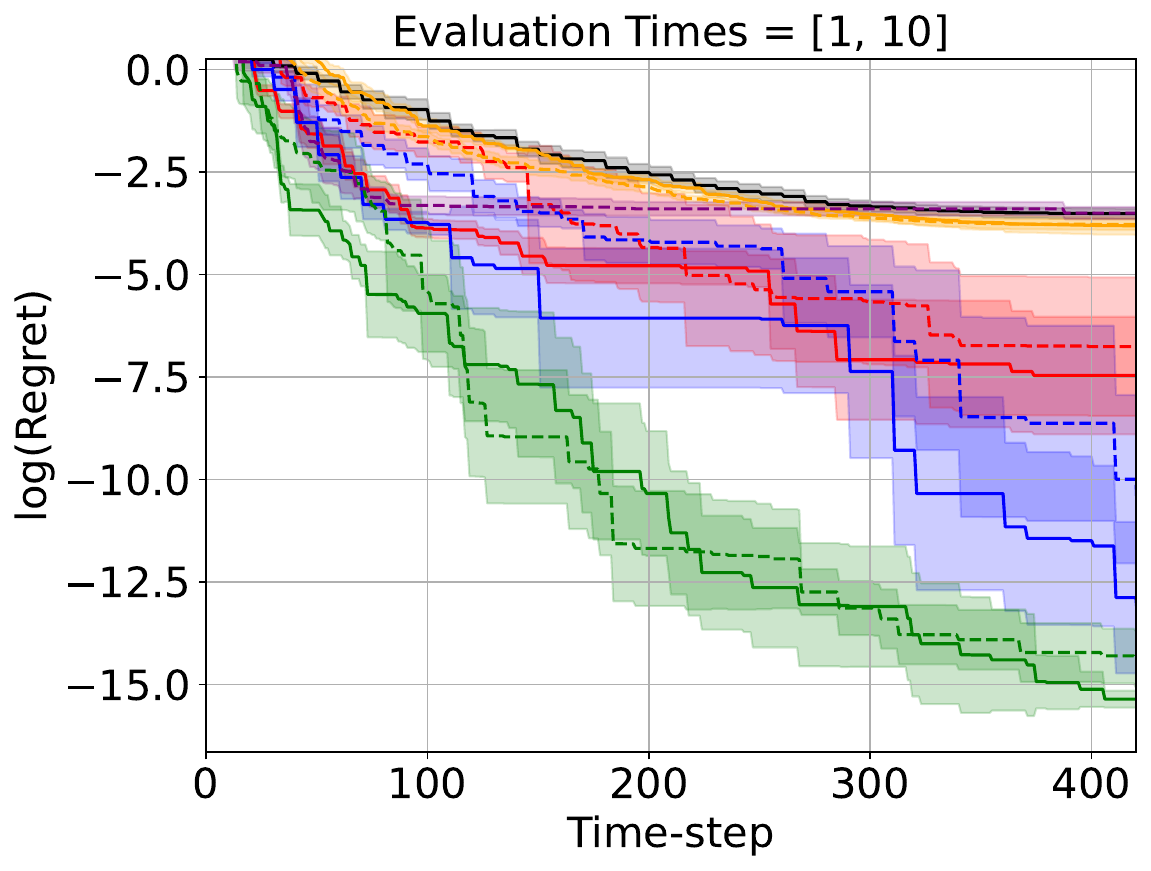}	
	\caption{Borehole8D}
	\end{subfigure}
	\caption{Results of synthetic experiments. We see that simply using UCB with our algorithm consistently achieves good results, always being one of the top performs. TuRBO also tends to perform fairly well, especially strongly in Hartmann6D, however, it also struggles badly in some benchmarks (Park4D and Borehole8D). MF-MES is not present in the Hartmann experiments due to the computational expense of testing the function when there is three fidelities.}
	\label{fig: mf_synthetic_results}
\end{figure}

In the interest of giving a fairer comparison, all single-query methods are turned into batch methods by selecting the first query using the acquisition function, and then filling the rest of the batch with random queries. We carry out the optimization until there is enough budget to get 200 high fidelity observations. For all experiments we set the batch size to 4, and assume that the batch space of the low and high fidelities is the same. The results can be see in Figure \ref{fig: mf_synthetic_results}, with the evaluation times for each fidelity shown in the title.

The Currin2D benchmark, is a simple function with a unique optima. Methods that use independent GPs for modelling perform badly. The strongest performers are the multi-fidelity versions of TuRBO (in yellow). In particular, we see that the best three performers (multi-fidelity TuRBO, and UCB-V-LP (in green)) achieve better regret (on average) after 150 time-steps than the rest of the methods in 400 -- they seem to leverage the lower fidelity to explore the space quickly and obtain and they begin to exploit the optimum quickly.

For the BadCurrin2D benchmark, the lower fidelity is simply the negative of the target fidelity, giving the worst possible approximation. However, methods using multi-output GPs (in yellow, green, purple) quickly learn the inverse correlation structure and leverage it to achieve basically identical results to the original benchmark. Methods using independent GPs (in red) on the other hand, have poor initial performance, but seem to recover well enough and have similar performance as in the original benchmark for larger budgets.

Hartmann3D is a smooth function with 4 local minima. TuRBO \textit{without} multi-fidelity (in black) seems to get stuck in one of these minima, however once we introduce the ability to quickly explore using cheap approximations, the performance of the method increases significantly (in yellow) since sub-optimal areas are discarded without centering the trust region in a local optima. We still observe larger uncertainties than for most methods, suggesting that getting stuck in local optima is still a possibility. On the other hand, simple upper confidence bound methods with multi-output GPs (in green) seem to perform better and provide better reliability due to the ability to explore, even after a local optima has been found.

In Hartmann6D we have another smooth function but with 6 local minima. This time multi-fidelity TuRBO methods (in yellow) perform much stronger than every other method. They achieve significantly lower regret and with very small uncertainty bounds. We conjecture this is because the \textit{initial} trust region is always large enough to contain the global optima (the size of the region depends on the initial estimated length-scales), and using the multi-fidelity approximations, the trust region can quickly center around the global optima. We further note that our modifications of UCB (in green), while far behind the multi-fidelity versions of TuRBO, they still outperform all the other benchmarks.

For Park4D and Borehole8D we obtain very similar results. In both benchmarks it seems that all versions of TuRBO (in yellow and black) perform very poorly. We conjecture this happens because the trust regions are initialized too small, leading to model misspecification (as all the samples lie in a small region), and therefore the algorithm getting stuck in local optima every time. On the other hand, in both cases we see very strong performance from our modified version of the UCB function (in green). The functions are simple enough, and the dimensionality low enough, so that exploration can be done quickly and effectively using the lower fidelity. The advantage of using multi-output GPs to transfer fidelity information is clear, as the independent GPs (in red) have significantly worse performance in both benchmarks.

Overall, we see very strong performance from at least one of the methods introduced in this work in each benchmark. Particularly strong are the methods that use multi-output GPs. We see little to no difference in performance when using the variance based or when using the information based criteria for choosing which fidelity to query. However, we recall that the information based method has less hyper-parameters, and note particularly strong performance in Hartmann6D.

\subsection{High-dimensional Optimization} \label{sec: exp_high_dim}
To highlight the usefulness of our proposed method, we show how we can use a specialist function to do high-dimensional multi-fidelity Bayesian Optimization. BO traditionally struggles in high-dimensions where the size of the search spaces means the exploratory phase of classical algorithms is given too much importance. \cite{eriksson2019scalable} proposed TuRBO which focuses on Local Optimization and tends to perform very well in high-dimensional settings. TuRBO uses Thompson Sampling to select the next experimental design, however, the experimental region is restricted to a trust region. The trust region is defined as a hyper-rectangle centered around the best experiment we have observed so far. If there are a lot of consecutive improvements (i.e. we consecutively obtain observations better than anything we had observed before), then we double the size of the region (i.e. we increase exploration), if there are too many consecutive failures, we shrink the size of the region (i.e. we increase exploitation). We use BoTorch's \citep{balandat2020botorch} implementation of TuRBO with the default hyper-parameters -- for changing the size of the region we only consider evaluation at the highest fidelity. \\

\begin{figure}[ht]
    \centering
	\includegraphics[width = 0.6\textwidth]{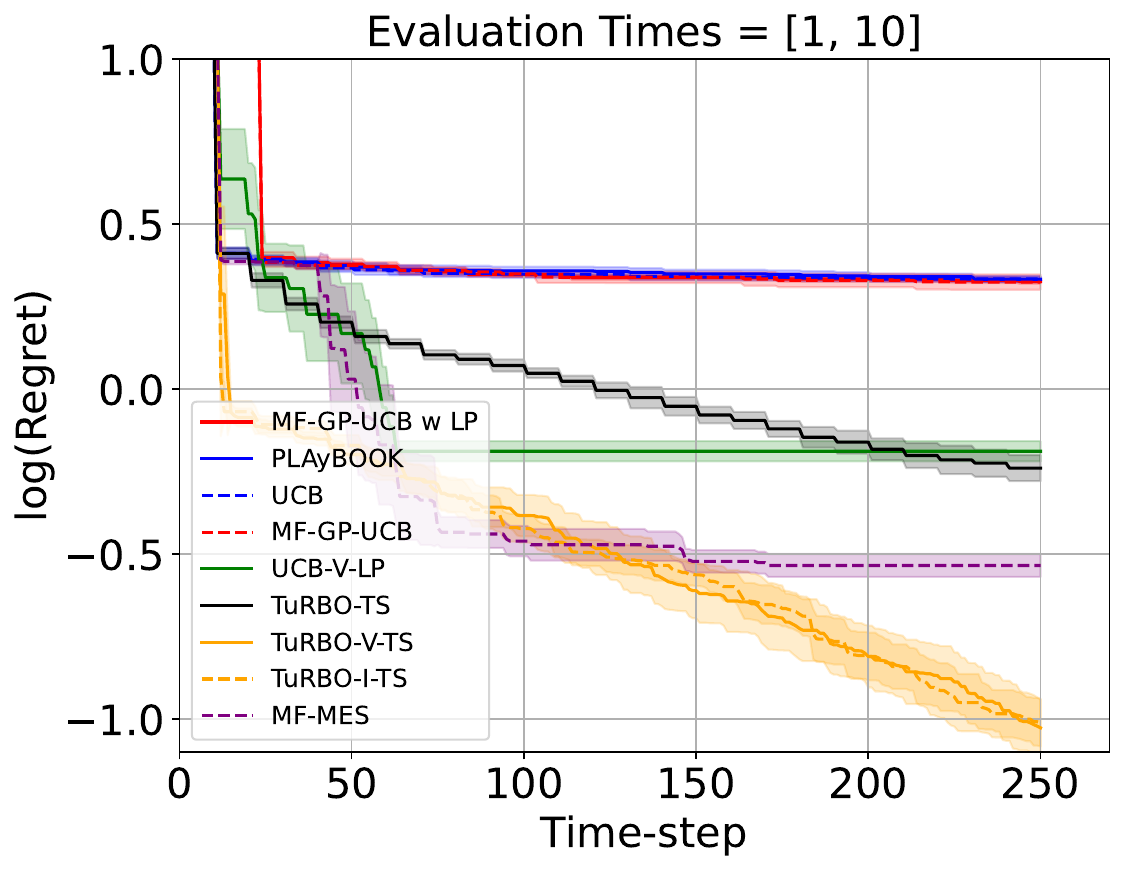}
	\caption{Results on Ackley 40D function. We can see that TuRBO is constantly improving, and outperforms other functions considerably. MF-MES and UCB-V-LP have good initial performance but begin to struggle after due to the over-exploration of classical BO.}
	\label{fig: mf_high_dim_results}
\end{figure}

Trust region approaches often work well on high-dimensional chemical engineering applications \cite{bajaj2018trust,kazi2021trust,kazi2022new}, since they avoid the problem of over-exploration in high-dimensional spaces. We show, using the Ackley 40D benchmark, that we can use this acquisition function in our algorithm to obtain much better performance than other benchmarks. For this case we set the batch size to 20, and again assume that the batch space of low and high fidelities is the same. We set the budget as to allow 500 high-fidelity observations. The results are shown in Figure \ref{fig: mf_high_dim_results}.

After less than 50 time-steps multi-fidelity TuRBO (in yellow) achieves better regret than PLAyBOOK (non-dashed blue), normal UCB (dashed blue) and MF-GP-UCB (both in red) achieve in the whole optimization, and similar performance to UCB-V-LP (green) and normal TuRBO (black). MF-MES also achieves very strong performance, however, it is not able to improve at the same rate that TuRBO does. All these results are expected, as the methods will over-explore in the high-dimensional space and never be able to exploit. TuRBO on the other hand, after a brief exploration phase, will shrink the trust region to something manageable and exploit locally, achieving much better regret. In addition we observe relatively small uncertainty regions, this due to the Ackley function being symmetric, and the high dimensionality of the problem making the starting points less likely to be relevant. The difference between the black and yellow lines gives strong evidence of the benefits of using multi-fidelity methods. We note, again, that there is little difference between the variance and information based multi-fidelity criteria (see both yellow lines).

\subsection{Battery Material Design}

\begin{figure}[ht]
    \centering
	\includegraphics[width = 0.6\textwidth]{ 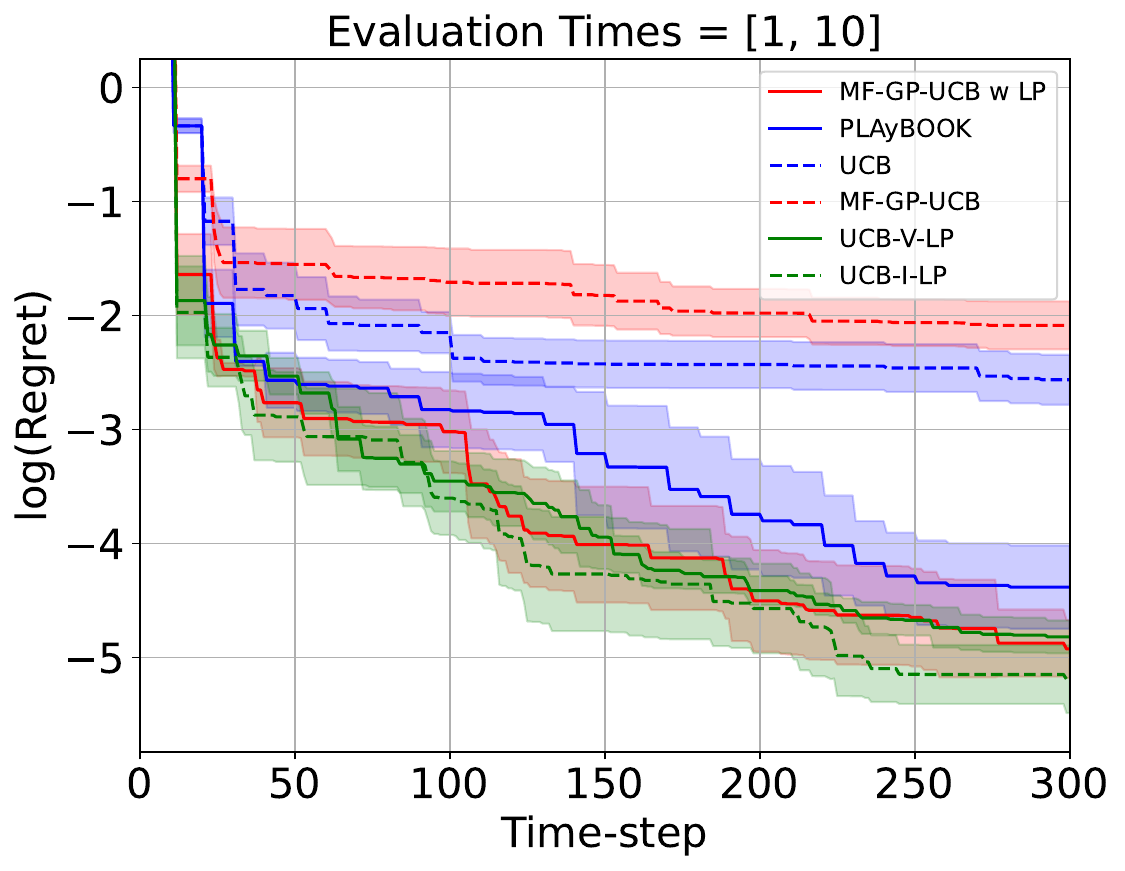}
	\caption{Results on battery design experiment. We can see that the asynchronous batching and multi-fidelity methods are significantly outperforming simple baselines (UCB), multi-fidelity without batching (MF-GP-UCB), and single-fidelity methods (PLAyBOOK). The performance of all batching multi-fidelity methods is similarly strong.}
	\label{fig: battery_results}
\end{figure}

\begin{figure}[ht]
    \centering
	\includegraphics[width = 0.8\textwidth]{ 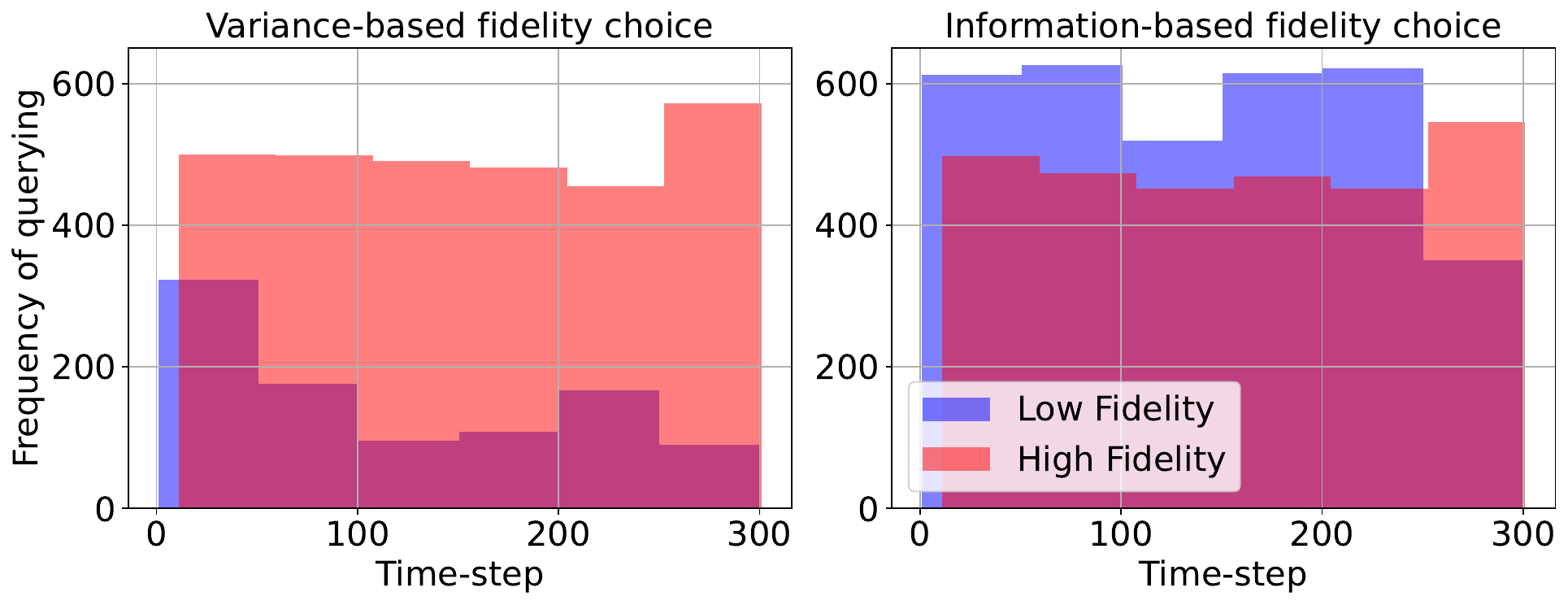}
	\caption{Experimental querying frequency comparison. The histogram above shows how often we queried the low and high fidelities for different time-intervals. We compare the querying frequency for the methods using UCB and Local Penalization, with different ways of choosing the fidelity -- variance-based (left) against information-based (right). The information-based fidelity criteria chooses to query the lower-fidelity more frequently, which suggests it could be the reason why it slightly outperforms the variance-based criteria.}
	\label{fig: querying_freq}
\end{figure}

We explore the application to the automatic design of battery materials. In this particular example, we aim to learn the best ratios of dopants in high-nickel cathode materials such that the specific capacity (electrochemically accessible charge per unit mass) is maximized. Our target is to find materials that have a high capacity over many charge-discharge cycles in pouch cell batteries. However, designing, building and testing such batteries requires many resources and long experimental cycles that take multiple months to process. As such, it is common for experiments to be carried out in parallel, where the batch size ranges between 10 and 20 concurrent queries, or experiments.

It is possible to obtain an approximation of pouch cell performance by considering coin cell batteries, where the materials are put into much smaller cylindrical containers requiring less material. In addition, we can reduce the duration of the experimental tests, perhaps by increasing the cycling rates or performing fewer charge-discharge cycles in total. Combining coin cell tests with accelerated test procedures allows us to get data in only a few weeks, and to carry out more experiments due to the lower material costs -- however some of the real problem's mechanisms may be suppressed. This represents the lower fidelity that we are interested in using.

For this example, we will be using an anonymized real-world data-set provided by BASF SE (the dataset, alongside the code used in the experiment section, can be found at \url{https://github.com/jpfolch/MFBoom}). We fit a Gaussian Process model to the data and using the method described in \cite{wilson2020efficiently}, we take a single sample to create our objective, $\tilde{f}^{(0)}$. We also took a second sample, $\tilde{f}^{(1)}$, to create the lower fidelity, meaning the function that we query is defined as:
\begin{equation}
f^{(m)}(x) = 
\begin{cases}
\tilde{f}^{(0)}(x) &\text{if } m = 0 \\
\tilde{f}^{(1)}(x) &\text{if } m = 1
\end{cases}
\end{equation}

Real battery materials laboratories vary in the degree of automation and the achievable experimental throughput. Here we assume we have a maximum batch budget of $20$ and consider differing batch spaces. We assume that for each coin cell battery, or low fidelity, querying costs 1 unit, while the pouch cell battery, or high fidelity, costs 2 units. This means we can concurrently do 20 coin cell experiments, or 10 pouch cell experiments, or a mix in between. We further assume that pouch cell experiments take 10 times longer than coin cell experiments, and set the time-budget to 300 (equivalent to being able to carry out 300 high-fidelity experiments). This is due to the combination of an accelerated test procedure with coin cell testing rather than an inherent speed-up associated with coin cells. The results are shown on Figure \ref{fig: battery_results}. 

We can see that the benefit to using the lower-fidelity with batching are substantial, as both methods that use both batching and multi-fidelity (green and non-dashed red) are significantly outperforming simple baselines, single-fidelity methods, and non-batching counterparts. For this specific example, we notice that using independent Gaussian Processes works surprisingly well, suggesting that the bias assumption is well suited to this benchmark. We did not compare against MF-MES or TuRBO due to complications in optimizing the acquisition function under the input constraints of the problem (see Appendix).

We further include a histogram, that shows how often we queried the low fidelity against the high fidelity. We compare UCB-V-LP (variance-based fidelity choice) against UCB-I-LP (information-based fidelity choice). The comparison can be see in Figure \ref{fig: querying_freq}. In both cases, we can see that towards the end of the optimization, there is decrease in low-fidelity querying, and an increase in high fidelity querying. However, information-based fidelity choice queries far more at the lowest fidelity -- and seems to get marginally better results. This highlights the advantage of the information-based method being parameter-free, as the variance-based requires better tuning of the UCB thresholds to use the lower-fidelity at its full capacity.

\section{Conclusion and Discussion}

In this paper we have explored methods in multi-fidelity and batch Bayesian Optimization, explaining why there is a close link between them and providing a comprehensive overview of the current state-of-the-art methods. We further show how we can leverage recent advances and propose an algorithm that allows practitioners to use acquisition functions well suited for their particular problem. We then empirically show the benefits of using such methods, illustrating their usefulness in both synthetic benchmarks, and real-life inspired problems.

Being able to choose the right acquisition function is of great importance in Bayesian Optimization. How to choose the correct acquisition function remains an open problem, and is far too broad a subject to tackle in this paper. However, it is clear that having flexibility in the choice of acquisition function is important. As an example, as shown in Figure \ref{fig: sub_hartmann6d}, the choice of TuRBO as an acquisition function leads to much better results than the others.

In some cases there can be clear advantages to using an acquisition function that is known to work well. As an example, consider the high dimensional optimization problem we showcase in section \ref{sec: exp_high_dim}, where TuRBO is known to perform well. However, no single acquisition function will perform optimally in every situation. Indeed, TuRBO performs very poorly in Park4D and Borehole8D where the single trust region gets stuck in a local optima. This highlights a second important benefit of being able to use any acquisition function: sometimes we should prefer simple solutions.

Complex methods such as TuRBO can provide strong state-of-the-art performance in many benchmarks, but generalize poorly unless their parameters are carefully tuned. The size and location of the initial trust region, how frequently to shrink or enlarge it, or the number of trust regions are examples of unintuitive choices that we must make when using the method. In contrast, once we have selected methods that fit the problem parameters (in this case multi-fidelity and batching), many practitioners and experimentalists will prefer to use a simple acquisition function as it is more interpretable, familiar, and choosing the hyper-parameters tends to be easier. Experimentally, we showed that using the UCB and local penalization (green lines in Figures \ref{fig: mf_synthetic_results}, \ref{fig: mf_high_dim_results}, and \ref{fig: battery_results}) always achieves a top four performance, regardless of the benchmark -- following Occam's Razor, we may choose the simplest model that fits the problem parameters.

With this work we hope to give readers a solid understanding of how to improve optimization procedures when you can parallelize experiments and obtain cheap approximations. We hope this will allow for more widespread use of such methods, because they exist and we show they can be incredibly useful. Indeed, it may be the case that many optimization procedures can benefit from multi-fidelity approximations but they have never been considered, or they have been thought of as something to do instead of parallelization, instead of doing them concurrently. We also hope that by making the algorithm flexible, it will appeal to a wide range of practitioners and problems.

\section*{Acknowledgments \& Disclosure of Funding}
JPF is funded by EPSRC through the Modern Statistics and Statistical Machine Learning (StatML) CDT (grant no. EP/S023151/1) and by BASF SE, Ludwigshafen am Rhein. The research was funded by Engineering \& Physical Sciences Research Council (EPSRC) Fellowships to RM and CT (grant no. EP/P016871/1 and EP/T001577/1). CT also acknowledges support from an Imperial College Research Fellowship.

\FloatBarrier

\bibliographystyle{unsrtnat}
\bibliography{main_arxiv.bib}  

\begin{thebibliography}{86}
\providecommand{\natexlab}[1]{#1}
\providecommand{\url}[1]{\texttt{#1}}
\expandafter\ifx\csname urlstyle\endcsname\relax
  \providecommand{\doi}[1]{doi: #1}\else
  \providecommand{\doi}{doi: \begingroup \urlstyle{rm}\Url}\fi

\bibitem[Chen et~al.(2019)Chen, Niu, Lee, Li, Yu, Xu, Zhang, Dufek,
  Whittingham, Meng, Xiao, and Liu]{CHEN20191094}
Shuru Chen, Chaojiang Niu, Hongkyung Lee, Qiuyan Li, Lu~Yu, Wu~Xu, Ji-Guang
  Zhang, Eric~J. Dufek, M.~Stanley Whittingham, Shirley Meng, Jie Xiao, and Jun
  Liu.
\newblock {Critical Parameters for Evaluating Coin Cells and Pouch Cells of
  Rechargeable Li-Metal Batteries}.
\newblock \emph{Joule}, 3\penalty0 (4):\penalty0 1094--1105, 2019.

\bibitem[Dörfler et~al.(2020)Dörfler, Althues, Härtel, Abendroth, Schumm,
  and Kaskel]{DORFLER2020539}
Susanne Dörfler, Holger Althues, Paul Härtel, Thomas Abendroth, Benjamin
  Schumm, and Stefan Kaskel.
\newblock {Challenges and Key Parameters of Lithium-Sulfur Batteries on Pouch
  Cell Level}.
\newblock \emph{Joule}, 4\penalty0 (3):\penalty0 539--554, 2020.

\bibitem[Liu et~al.(2021)Liu, Liu, Li, and Gao]{liu2021strategy}
Ya-Tao Liu, Sheng Liu, Guo-Ran Li, and Xue-Ping Gao.
\newblock Strategy of enhancing the volumetric energy density for
  lithium--sulfur batteries.
\newblock \emph{Advanced Materials}, 33\penalty0 (8):\penalty0 2003955, 2021.

\bibitem[Attia et~al.(2020)Attia, Grover, Jin, Severson, Markov, Liao, Chen,
  Cheong, Perkins, Yang, et~al.]{attia2020closed}
Peter~M Attia, Aditya Grover, Norman Jin, Kristen~A Severson, Todor~M Markov,
  Yang-Hung Liao, Michael~H Chen, Bryan Cheong, Nicholas Perkins, Zi~Yang,
  et~al.
\newblock Closed-loop optimization of fast-charging protocols for batteries
  with machine learning.
\newblock \emph{Nature}, 578\penalty0 (7795):\penalty0 397--402, 2020.

\bibitem[Asprey and Macchietto(2000)]{asprey2000statistical}
SP~Asprey and S~Macchietto.
\newblock Statistical tools for optimal dynamic model building.
\newblock \emph{Computers \& Chemical Engineering}, 24\penalty0 (2-7):\penalty0
  1261--1267, 2000.

\bibitem[Box and Lucas(1959)]{box1959design}
George~EP Box and HL~Lucas.
\newblock Design of experiments in non-linear situations.
\newblock \emph{Biometrika}, 46\penalty0 (1/2):\penalty0 77--90, 1959.

\bibitem[Waldron et~al.(2019)Waldron, Pankajakshan, Quaglio, Cao, Galvanin, and
  Gavriilidis]{waldron2019autonomous}
Conor Waldron, Arun Pankajakshan, Marco Quaglio, Enhong Cao, Federico Galvanin,
  and Asterios Gavriilidis.
\newblock An autonomous microreactor platform for the rapid identification of
  kinetic models.
\newblock \emph{Reaction Chemistry \& Engineering}, 4:\penalty0 1623--1636,
  2019.

\bibitem[Hunter and Reiner(1965)]{hunter1965designs}
William~G Hunter and Albey~M Reiner.
\newblock Designs for discriminating between two rival models.
\newblock \emph{Technometrics}, 7\penalty0 (3):\penalty0 307--323, 1965.

\bibitem[Tsay et~al.(2017)Tsay, Pattison, Baldea, Weinstein, Hodson, and
  Johnson]{tsay2017doeforparameterestimation}
Calvin Tsay, Richard~C. Pattison, Michael Baldea, Ben Weinstein, Steven~J.
  Hodson, and Robert~D. Johnson.
\newblock A superstructure-based design of experiments framework for
  simultaneous domain-restricted model identification and parameter estimation.
\newblock \emph{Computers \& Chemical Engineering}, 107:\penalty0 408--426,
  2017.

\bibitem[Vincent(2016)]{vincent2016hierarchical}
Benjamin~T Vincent.
\newblock {Hierarchical Bayesian estimation and hypothesis testing for delay
  discounting tasks}.
\newblock \emph{Behavior Research Methods}, 48\penalty0 (4):\penalty0
  1608--1620, 2016.

\bibitem[Foster et~al.(2021)Foster, Ivanova, Malik, and
  Rainforth]{foster2021deep}
Adam Foster, Desi~R Ivanova, Ilyas Malik, and Tom Rainforth.
\newblock {Deep Adaptive Design: Amortizing Sequential Bayesian Experimental
  Design}.
\newblock In \emph{International Conference on Machine Learning}, pages
  3384--3395, 2021.

\bibitem[Franceschini and Macchietto(2008)]{franceschini2008model}
Gaia Franceschini and Sandro Macchietto.
\newblock Model-based design of experiments for parameter precision: State of
  the art.
\newblock \emph{Chemical Engineering Science}, 63\penalty0 (19):\penalty0
  4846--4872, 2008.

\bibitem[Wang and Dowling(2022{\natexlab{a}})]{wang2022pyomo}
Jialu Wang and Alexander~W Dowling.
\newblock {Pyomo.DOE: An open-source package for model-based design of
  experiments in Python}.
\newblock \emph{AIChE Journal}, page e17813, 2022{\natexlab{a}}.

\bibitem[Bajaj et~al.(2021)Bajaj, Arora, and Hasan]{Bajaj2021blackbox}
Ishan Bajaj, Akhil Arora, and M.~M.~Faruque Hasan.
\newblock \emph{Black-Box Optimization: Methods and Applications}, pages
  35--65.
\newblock Springer International Publishing, 2021.

\bibitem[Jones et~al.(1998)Jones, Schonlau, and Welch]{jones1998efficient}
Donald~R Jones, Matthias Schonlau, and William~J Welch.
\newblock Efficient global optimization of expensive black-box functions.
\newblock \emph{Journal of Global Optimization}, 13\penalty0 (4):\penalty0
  455--492, 1998.

\bibitem[Shahriari et~al.(2016)Shahriari, Swersky, Wang, Adams, and
  de~Freitas]{shahriari2016bo}
Bobak Shahriari, Kevin Swersky, Ziyu Wang, Ryan~P. Adams, and Nando de~Freitas.
\newblock {Taking the Human Out of the Loop: A Review of Bayesian
  Optimization}.
\newblock \emph{Proceedings of the IEEE}, 104\penalty0 (1):\penalty0 148--175,
  2016.

\bibitem[Bhosekar and Ierapetritou(2018)]{bhosekar2018advances}
Atharv Bhosekar and Marianthi Ierapetritou.
\newblock Advances in surrogate based modeling, feasibility analysis, and
  optimization: {A} review.
\newblock \emph{Computers \& Chemical Engineering}, 108:\penalty0 250--267,
  2018.

\bibitem[Thebelt et~al.(2022{\natexlab{a}})Thebelt, Wiebe, Kronqvist, Tsay, and
  Misener]{thebelt2022maximizingcheminfo}
Alexander Thebelt, Johannes Wiebe, Jan Kronqvist, Calvin Tsay, and Ruth
  Misener.
\newblock Maximizing information from chemical engineering data sets:
  {A}pplications to machine learning.
\newblock \emph{Chemical Engineering Science}, 252:\penalty0 117469,
  2022{\natexlab{a}}.

\bibitem[Kandasamy et~al.(2016{\natexlab{a}})Kandasamy, Dasarathy, Oliva,
  Schneider, and P{\'o}czos]{kandasamy2019multi}
Kirthevasan Kandasamy, Gautam Dasarathy, Junier~B Oliva, Jeff Schneider, and
  Barnab{\'a}s P{\'o}czos.
\newblock {Gaussian Process Bandit Optimisation with Multi-fidelity
  Evaluations}.
\newblock \emph{Advances in Neural Information Processing Systems}, 29,
  2016{\natexlab{a}}.

\bibitem[Takeno et~al.(2020)Takeno, Fukuoka, Tsukada, Koyama, Shiga, Takeuchi,
  and Karasuyama]{takeno2020multi}
Shion Takeno, Hitoshi Fukuoka, Yuhki Tsukada, Toshiyuki Koyama, Motoki Shiga,
  Ichiro Takeuchi, and Masayuki Karasuyama.
\newblock Multi-fidelity {Bayesian} optimization with max-value entropy search
  and its parallelization.
\newblock In \emph{International Conference on Machine Learning}, pages
  9334--9345. PMLR, 2020.

\bibitem[Gonz{\'a}lez et~al.(2016)Gonz{\'a}lez, Dai, Hennig, and
  Lawrence]{gonzalez2016batch}
Javier Gonz{\'a}lez, Zhenwen Dai, Philipp Hennig, and Neil Lawrence.
\newblock {Batch Bayesian Optimization via Local Penalization}.
\newblock In \emph{Proceedings of the 19th International Conference on
  Artificial Intelligence and Statistics}, pages 648--657, 09--11 May 2016.

\bibitem[Alvi et~al.(2019)Alvi, Ru, Calliess, Roberts, and
  Osborne]{alvi2019localpen}
Ahsan Alvi, Binxin Ru, Jan-Peter Calliess, Stephen Roberts, and Michael~A.
  Osborne.
\newblock {Asynchronous Batch Bayesian Optimisation with Improved Local
  Penalisation}.
\newblock In \emph{International Conference on Machine Learning}, pages
  253--262, 09--15 Jun 2019.

\bibitem[Kandasamy et~al.(2018)Kandasamy, Krishnamurthy, Schneider, and
  Poczos]{pmlr-v84-kandasamy18a}
Kirthevasan Kandasamy, Akshay Krishnamurthy, Jeff Schneider, and Barnabas
  Poczos.
\newblock Parallelised {Bayesian Optimisation} via {Thompson Sampling}.
\newblock In \emph{Proceedings of the Twenty-First International Conference on
  Artificial Intelligence and Statistics}, pages 133--142, 2018.

\bibitem[Snoek et~al.(2012)Snoek, Larochelle, and Adams]{snoek2012practical}
Jasper Snoek, Hugo Larochelle, and Ryan~P Adams.
\newblock Practical {Bayesian} optimization of machine learning algorithms.
\newblock \emph{Advances in Neural Information Processing Systems}, 25, 2012.

\bibitem[Bergstra et~al.(2011)Bergstra, Bardenet, Bengio, and
  K{\'e}gl]{bergstra2011algorithms}
James Bergstra, R{\'e}mi Bardenet, Yoshua Bengio, and Bal{\'a}zs K{\'e}gl.
\newblock Algorithms for hyper-parameter optimization.
\newblock \emph{Advances in Neural Information Processing Systems}, 24, 2011.

\bibitem[Li et~al.(2017)Li, Jamieson, DeSalvo, Rostamizadeh, and
  Talwalkar]{li2017hyperband}
Lisha Li, Kevin Jamieson, Giulia DeSalvo, Afshin Rostamizadeh, and Ameet
  Talwalkar.
\newblock Hyperband: A novel bandit-based approach to hyperparameter
  optimization.
\newblock \emph{The Journal of Machine Learning Research}, 18\penalty0
  (1):\penalty0 6765--6816, 2017.

\bibitem[Amaran et~al.(2016)Amaran, Sahinidis, Sharda, and
  Bury]{amaran2016simulation}
Satyajith Amaran, Nikolaos~V Sahinidis, Bikram Sharda, and Scott~J Bury.
\newblock Simulation optimization: a review of algorithms and applications.
\newblock \emph{Annals of Operations Research}, 240\penalty0 (1):\penalty0
  351--380, 2016.

\bibitem[Wang and Dowling(2022{\natexlab{b}})]{WANG2022100728}
Ke~Wang and Alexander~W Dowling.
\newblock Bayesian optimization for chemical products and functional materials.
\newblock \emph{Current Opinion in Chemical Engineering}, 36:\penalty0 100728,
  2022{\natexlab{b}}.

\bibitem[Folch et~al.(2022)Folch, Zhang, Lee, Shafei, Walz, Tsay, van~der Wilk,
  and Misener]{folch2022snake}
Jose~Pablo Folch, Shiqiang Zhang, Robert~M Lee, Behrang Shafei, David Walz,
  Calvin Tsay, Mark van~der Wilk, and Ruth Misener.
\newblock {SnAKe}: {Bayesian} {Optimization} with {Pathwise} {Exploration}.
\newblock \emph{arXiv preprint arXiv:2202.00060}, 2022.

\bibitem[Thebelt et~al.(2021)Thebelt, Kronqvist, Mistry, Lee, Sudermann-Merx,
  and Misener]{thebelt2021entmoot}
Alexander Thebelt, Jan Kronqvist, Miten Mistry, Robert~M Lee, Nathan
  Sudermann-Merx, and Ruth Misener.
\newblock {ENTMOOT}: a framework for optimization over ensemble tree models.
\newblock \emph{Computers \& Chemical Engineering}, 151:\penalty0 107343, 2021.

\bibitem[Thebelt et~al.(2022{\natexlab{b}})Thebelt, Tsay, Lee, Sudermann-Merx,
  Walz, Tranter, and Misener]{thebelt2022multi}
Alexander Thebelt, Calvin Tsay, Robert~M Lee, Nathan Sudermann-Merx, David
  Walz, Tom Tranter, and Ruth Misener.
\newblock Multi-objective constrained optimization for energy applications via
  tree ensembles.
\newblock \emph{Applied Energy}, 306:\penalty0 118061, 2022{\natexlab{b}}.

\bibitem[Thebelt et~al.(2022{\natexlab{c}})Thebelt, Tsay, Lee, Sudermann-Merx,
  Walz, Shafei, and Misener]{thebelt2022tree}
Alexander Thebelt, Calvin Tsay, Robert~M Lee, Nathan Sudermann-Merx, David
  Walz, Behrang Shafei, and Ruth Misener.
\newblock Tree ensemble kernels for {B}ayesian optimization with known
  constraints over mixed-feature spaces.
\newblock \emph{arXiv preprint arXiv:2207.00879}, 2022{\natexlab{c}}.

\bibitem[Cozad et~al.(2014)Cozad, Sahinidis, and Miller]{cozad2014learning}
Alison Cozad, Nikolaos~V Sahinidis, and David~C Miller.
\newblock Learning surrogate models for simulation-based optimization.
\newblock \emph{AIChE Journal}, 60\penalty0 (6):\penalty0 2211--2227, 2014.

\bibitem[Boukouvala and Floudas(2017)]{boukouvala2017argonaut}
Fani Boukouvala and Christodoulos~A Floudas.
\newblock {ARGONAUT: AlgoRithms for Global Optimization of coNstrAined grey-box
  compUTational problems}.
\newblock \emph{Optimization Letters}, 11\penalty0 (5):\penalty0 895--913,
  2017.

\bibitem[Olofsson et~al.(2018)Olofsson, Mehrian, Calandra, Geris, Deisenroth,
  and Misener]{olofsson2018bayesian}
Simon Olofsson, Mohammad Mehrian, Roberto Calandra, Liesbet Geris, Marc~Peter
  Deisenroth, and Ruth Misener.
\newblock Bayesian multiobjective optimisation with mixed analytical and
  black-box functions: Application to tissue engineering.
\newblock \emph{IEEE Transactions on Biomedical Engineering}, 66\penalty0
  (3):\penalty0 727--739, 2018.

\bibitem[Park et~al.(2018)Park, Na, Kim, and Lee]{park2018multi}
Seongeon Park, Jonggeol Na, Minjun Kim, and Jong~Min Lee.
\newblock {Multi-objective Bayesian optimization of chemical reactor design
  using computational fluid dynamics}.
\newblock \emph{Computers \& Chemical Engineering}, 119:\penalty0 25--37, 2018.

\bibitem[Paulson and Lu(2022)]{paulson2022cobalt}
Joel~A Paulson and Congwen Lu.
\newblock {COBALT: COnstrained Bayesian optimizAtion of computationaLly
  expensive grey-box models exploiting derivaTive information}.
\newblock \emph{Computers \& Chemical Engineering}, 160:\penalty0 107700, 2022.

\bibitem[Kudva et~al.(2022)Kudva, Sorourifar, and
  Paulson]{kudva2022constrained}
Akshay Kudva, Farshud Sorourifar, and Joel~A Paulson.
\newblock {Constrained robust Bayesian optimization of expensive noisy
  black-box functions with guaranteed regret bounds}.
\newblock \emph{AIChE Journal}, page e17857, 2022.

\bibitem[Kennedy and O'Hagan(2000)]{kennedy2000multifidelity}
M.~C. Kennedy and A.~O'Hagan.
\newblock {Predicting the Output from a Complex Computer Code When Fast
  Approximations Are Available}.
\newblock \emph{Biometrika}, 87\penalty0 (1):\penalty0 1--13, 2000.

\bibitem[Gratiet(2013)]{gratiet2013recursive}
Loic~Le Gratiet.
\newblock {Recursive Co-Kriging Model for Design of Computer Experiments with
  Multiple Levels of Fidelity with an Application to Hydrodynamic}, 2013.

\bibitem[Cutajar et~al.(2019)Cutajar, Pullin, Damianou, Lawrence, and
  Gonzalez]{cutajar2019deep}
Kurt Cutajar, Mark Pullin, Andreas Damianou, Neil Lawrence, and Javier
  Gonzalez.
\newblock {Deep Gaussian Processes} for {Multi-fidelity Modeling}, 2019.

\bibitem[Kandasamy et~al.(2016{\natexlab{b}})Kandasamy, Dasarathy, Schneider,
  and Poczos]{kandasamy2016multifidelityKbandit}
Kirthevasan Kandasamy, Gautam Dasarathy, Jeff Schneider, and Barnabas Poczos.
\newblock The {Multi-fidelity Multi-armed Bandit}, 2016{\natexlab{b}}.

\bibitem[Sen et~al.(2018)Sen, Kandasamy, and Shakkottai]{sen18a}
Rajat Sen, Kirthevasan Kandasamy, and Sanjay Shakkottai.
\newblock {Multi-Fidelity Black-Box Optimization} with {Hierarchical
  Partitions}.
\newblock In \emph{International Conference on Machine Learning}, pages
  4538--4547, 2018.

\bibitem[Journel and Huijbregts(1976)]{journel1976mining}
Andre~G Journel and Charles~J Huijbregts.
\newblock Mining geostatistics.
\newblock 1976.

\bibitem[Goovaerts et~al.(1997)]{goovaerts1997geostatistics}
Pierre Goovaerts et~al.
\newblock \emph{Geostatistics for natural resources evaluation}.
\newblock Oxford University Press on Demand, 1997.

\bibitem[Alvarez et~al.(2012)Alvarez, Rosasco, Lawrence,
  et~al.]{alvarez2012kernels}
Mauricio~A Alvarez, Lorenzo Rosasco, Neil~D Lawrence, et~al.
\newblock Kernels for vector-valued functions: A review.
\newblock \emph{Foundations and Trends{\textregistered} in Machine Learning},
  4\penalty0 (3):\penalty0 195--266, 2012.

\bibitem[Zhuang et~al.(2021)Zhuang, Qi, Duan, Xi, Zhu, Zhu, Xiong, and
  He]{zhuang2021transferlearningsurvery}
Fuzhen Zhuang, Zhiyuan Qi, Keyu Duan, Dongbo Xi, Yongchun Zhu, Hengshu Zhu, Hui
  Xiong, and Qing He.
\newblock A comprehensive survey on transfer learning.
\newblock \emph{Proceedings of the IEEE}, 109\penalty0 (1):\penalty0 43--76,
  2021.

\bibitem[Pan and Yang(2009)]{pan2009survey}
Sinno~Jialin Pan and Qiang Yang.
\newblock A survey on transfer learning.
\newblock \emph{IEEE Transactions on knowledge and data engineering},
  22\penalty0 (10):\penalty0 1345--1359, 2009.

\bibitem[Li et~al.(2020)Li, Gu, Zhang, and Chen]{li2020transfer}
Weijun Li, Sai Gu, Xiangping Zhang, and Tao Chen.
\newblock Transfer learning for process fault diagnosis: Knowledge transfer
  from simulation to physical processes.
\newblock \emph{Computers \& Chemical Engineering}, 139:\penalty0 106904, 2020.

\bibitem[Li and Rangarajan(2022)]{li2022conceptual}
Bowen Li and Srinivas Rangarajan.
\newblock A conceptual study of transfer learning with linear models for
  data-driven property prediction.
\newblock \emph{Computers \& Chemical Engineering}, 157:\penalty0 107599, 2022.

\bibitem[Jia et~al.(2020)Jia, Zhang, and You]{jia2020transfer}
Runda Jia, Shulei Zhang, and Fengqi You.
\newblock Transfer learning for end-product quality prediction of batch
  processes using domain-adaption {joint-Y PLS}.
\newblock \emph{Computers \& Chemical Engineering}, 140:\penalty0 106943, 2020.

\bibitem[Rogers et~al.(2022)Rogers, Vega-Ramon, Yan, del R{\'\i}o-Chanona,
  Jing, and Zhang]{rogers2022transfer}
Alexander~W Rogers, Fernando Vega-Ramon, Jiangtao Yan, Ehecatl~A del
  R{\'\i}o-Chanona, Keju Jing, and Dongda Zhang.
\newblock A transfer learning approach for predictive modeling of bioprocesses
  using small data.
\newblock \emph{Biotechnology and Bioengineering}, 119\penalty0 (2):\penalty0
  411--422, 2022.

\bibitem[Ginsbourger et~al.(2011)Ginsbourger, Janusevskis, and
  Le~Riche]{ginsbourger2011dealing}
David Ginsbourger, Janis Janusevskis, and Rodolphe Le~Riche.
\newblock \emph{{Dealing with Asynchronicity in Parallel Gaussian Process based
  Global Optimization}}.
\newblock PhD thesis, Mines Saint-Etienne, 2011.

\bibitem[Alshehri et~al.(2020)Alshehri, Gani, and You]{alshehri2020deep}
Abdulelah~S Alshehri, Rafiqul Gani, and Fengqi You.
\newblock {Deep learning and knowledge-based methods for computer-aided
  molecular design—toward a unified approach: State-of-the-art and future
  directions}.
\newblock \emph{Computers \& Chemical Engineering}, 141:\penalty0 107005, 2020.

\bibitem[Coley(2021)]{coley2021defining}
Connor~W Coley.
\newblock Defining and exploring chemical spaces.
\newblock \emph{Trends in Chemistry}, 3\penalty0 (2):\penalty0 133--145, 2021.

\bibitem[Coley et~al.(2019)Coley, Thomas, Lummiss, Jaworski, Breen, Schultz,
  Hart, Fishman, Rogers, Gao, Hicklin, Plehiers, Byington, Piotti, Green, Hart,
  Jamison, and Jensen]{coley2019robotic}
Connor~W. Coley, Dale~A. Thomas, Justin A.~M. Lummiss, Jonathan~N. Jaworski,
  Christopher~P. Breen, Victor Schultz, Travis Hart, Joshua~S. Fishman, Luke
  Rogers, Hanyu Gao, Robert~W. Hicklin, Pieter~P. Plehiers, Joshua Byington,
  John~S. Piotti, William~H. Green, A.~John Hart, Timothy~F. Jamison, and
  Klavs~F. Jensen.
\newblock A robotic platform for flow synthesis of organic compounds informed
  by ai planning.
\newblock \emph{Science}, 365\penalty0 (6453), 2019.

\bibitem[Li et~al.(2021)Li, Kirby, and Zhe]{li2021batch}
Shibo Li, Robert Kirby, and Shandian Zhe.
\newblock Batch multi-fidelity {Bayesian} optimization with deep
  auto-regressive networks.
\newblock \emph{Advances in Neural Information Processing Systems}, 34, 2021.

\bibitem[Wang and Jegelka(2017)]{wang2017max}
Zi~Wang and Stefanie Jegelka.
\newblock Max-value entropy search for efficient {} optimization.
\newblock In \emph{International Conference on Machine Learning}, pages
  3627--3635. PMLR, 2017.

\bibitem[Moss et~al.(2021)Moss, Leslie, Gonzalez, and Rayson]{moss2021gibbon}
Henry~B Moss, David~S Leslie, Javier Gonzalez, and Paul Rayson.
\newblock {GIBBON}: General-purpose information-based {B}ayesian optimisation.
\newblock \emph{Journal of Machine Learning Research}, 22\penalty0
  (235):\penalty0 1--49, 2021.

\bibitem[H{\"u}llen et~al.(2020)H{\"u}llen, Zhai, Kim, Sinha, Realff, and
  Boukouvala]{hullen2020managing}
Gordon H{\"u}llen, Jianyuan Zhai, Sun~Hye Kim, Anshuman Sinha, Matthew~J
  Realff, and Fani Boukouvala.
\newblock Managing uncertainty in data-driven simulation-based optimization.
\newblock \emph{Computers \& Chemical Engineering}, 136:\penalty0 106519, 2020.

\bibitem[Rasmussen and Williams(2005)]{rasmussen2005gps}
Carl~Edward Rasmussen and Christopher K.~I. Williams.
\newblock \emph{Gaussian Processes for Machine Learning (Adaptive Computation
  and Machine Learning)}.
\newblock The MIT Press, 2005.

\bibitem[Teh et~al.(2005)Teh, Seeger, and Jordan]{teh2005semiparametric}
Yee~Whye Teh, Matthias Seeger, and Michael~I Jordan.
\newblock Semiparametric latent factor models.
\newblock In \emph{International Workshop on Artificial Intelligence and
  Statistics}, pages 333--340. PMLR, 2005.

\bibitem[Ver~Hoef and Barry(1998)]{ver1998constructing}
Jay~M Ver~Hoef and Ronald~Paul Barry.
\newblock Constructing and fitting models for cokriging and multivariable
  spatial prediction.
\newblock \emph{Journal of Statistical Planning and Inference}, 69\penalty0
  (2):\penalty0 275--294, 1998.

\bibitem[Higdon(2002)]{higdon2002space}
Dave Higdon.
\newblock Space and space-time modeling using process convolutions.
\newblock In \emph{Quantitative methods for current environmental issues},
  pages 37--56. Springer, 2002.

\bibitem[van~der Wilk et~al.(2017)van~der Wilk, Rasmussen, and
  Hensman]{van2017convolutional}
Mark van~der Wilk, Carl~Edward Rasmussen, and James Hensman.
\newblock {Convolutional Gaussian Processes}.
\newblock \emph{Advances in Neural Information Processing Systems}, 30, 2017.

\bibitem[Damianou and Lawrence(2013)]{damianou2013deep}
Andreas Damianou and Neil~D Lawrence.
\newblock Deep {Gaussian} processes.
\newblock In \emph{Artificial intelligence and statistics}, pages 207--215.
  PMLR, 2013.

\bibitem[MacKay(1992)]{mackay1992practical}
David~JC MacKay.
\newblock A practical {Bayesian} framework for backpropagation networks.
\newblock \emph{Neural Computation}, 4\penalty0 (3):\penalty0 448--472, 1992.

\bibitem[Neal(2012)]{neal2012bayesian}
Radford~M Neal.
\newblock \emph{Bayesian learning for neural networks}, volume 118.
\newblock Springer Science \& Business Media, 2012.

\bibitem[Izmailov et~al.(2021)Izmailov, Vikram, Hoffman, and
  Wilson]{izmailov2021bayesian}
Pavel Izmailov, Sharad Vikram, Matthew~D Hoffman, and Andrew Gordon~Gordon
  Wilson.
\newblock {What are Bayesian neural network posteriors really like?}
\newblock In \emph{International Conference on Machine Learning}, pages
  4629--4640, 2021.

\bibitem[Villemonteix et~al.(2009)Villemonteix, Vazquez, and
  Walter]{villemonteix2009informational}
Julien Villemonteix, Emmanuel Vazquez, and Eric Walter.
\newblock An informational approach to the global optimization of
  expensive-to-evaluate functions.
\newblock \emph{Journal of Global Optimization}, 44\penalty0 (4):\penalty0
  509--534, 2009.

\bibitem[Hennig and Schuler(2012)]{hennig2012entropy}
Philipp Hennig and Christian~J Schuler.
\newblock {Entropy Search for Information-Efficient Global Optimization}.
\newblock \emph{Journal of Machine Learning Research}, 13\penalty0 (6), 2012.

\bibitem[Hern{\'a}ndez-Lobato et~al.(2014)Hern{\'a}ndez-Lobato, Hoffman, and
  Ghahramani]{hernandez2014predictive}
Jos{\'e}~Miguel Hern{\'a}ndez-Lobato, Matthew~W Hoffman, and Zoubin Ghahramani.
\newblock Predictive entropy search for efficient global optimization of
  black-box functions.
\newblock \emph{Advances in Neural Information Processing Systems}, 27, 2014.

\bibitem[Tu et~al.(2022)Tu, Gandy, Kantas, and Shafei]{tu2022joint}
Ben Tu, Axel Gandy, Nikolas Kantas, and Behrang Shafei.
\newblock Joint entropy search for multi-objective {B}ayesian optimization.
\newblock \emph{arXiv preprint arXiv:2210.02905}, 2022.

\bibitem[Hvarfner et~al.(2022)Hvarfner, Hutter, and Nardi]{hvarfner2022joint}
Carl Hvarfner, Frank Hutter, and Luigi Nardi.
\newblock Joint entropy search for maximally-informed {B}ayesian optimization.
\newblock \emph{arXiv preprint arXiv:2206.04771}, 2022.

\bibitem[Schweidtmann et~al.(2018)Schweidtmann, Clayton, Holmes, Bradford,
  Bourne, and Lapkin]{schweidtmann2018machine}
Artur~M Schweidtmann, Adam~D Clayton, Nicholas Holmes, Eric Bradford, Richard~A
  Bourne, and Alexei~A Lapkin.
\newblock Machine learning meets continuous flow chemistry: {A}utomated
  optimization towards the {P}areto front of multiple objectives.
\newblock \emph{Chemical Engineering Journal}, 352:\penalty0 277--282, 2018.

\bibitem[Badejo and Ierapetritou(2022)]{badejo2022integrating}
Oluwadare Badejo and Marianthi Ierapetritou.
\newblock Integrating tactical planning, operational planning and scheduling
  using data-driven feasibility analysis.
\newblock \emph{Computers \& Chemical Engineering}, 161:\penalty0 107759, 2022.

\bibitem[Srinivas et~al.(2010)Srinivas, Krause, Kakade, and
  Seeger]{srinivas2009gaussian}
Niranjan Srinivas, Andreas Krause, Sham Kakade, and Matthias Seeger.
\newblock {Gaussian Process Optimization in the Bandit Setting: No Regret and
  Experimental Design}.
\newblock In \emph{International Conference on Machine Learning}, pages
  1015--1022, 2010.

\bibitem[Eriksson et~al.(2019)Eriksson, Pearce, Gardner, Turner, and
  Poloczek]{eriksson2019scalable}
David Eriksson, Michael Pearce, Jacob Gardner, Ryan~D Turner, and Matthias
  Poloczek.
\newblock Scalable global optimization via local {Bayesian} optimization.
\newblock \emph{Advances in Neural Information Processing Systems}, 32, 2019.

\bibitem[Paszke et~al.(2019)Paszke, Gross, Massa, Lerer, Bradbury, Chanan,
  Killeen, Lin, Gimelshein, Antiga, Desmaison, Kopf, Yang, DeVito, Raison,
  Tejani, Chilamkurthy, Steiner, Fang, Bai, and Chintala]{paszke2019pytorch}
Adam Paszke, Sam Gross, Francisco Massa, Adam Lerer, James Bradbury, Gregory
  Chanan, Trevor Killeen, Zeming Lin, Natalia Gimelshein, Luca Antiga, Alban
  Desmaison, Andreas Kopf, Edward Yang, Zachary DeVito, Martin Raison, Alykhan
  Tejani, Sasank Chilamkurthy, Benoit Steiner, Lu~Fang, Junjie Bai, and Soumith
  Chintala.
\newblock {PyTorch: An Imperative Style, High-Performance Deep Learning
  Library}.
\newblock In \emph{Advances in Neural Information Processing Systems}, pages
  8024--8035. Curran Associates, Inc., 2019.

\bibitem[Gardner et~al.(2018)Gardner, Pleiss, Bindel, Weinberger, and
  Wilson]{gardner2018gpytorch}
{Jacob R.} Gardner, Geoff Pleiss, David Bindel, {Kilian Q.} Weinberger, and
  {Andrew Gordon} Wilson.
\newblock {Gpytorch: Blackbox matrix-matrix Gaussian Process inference with GPU
  acceleration}.
\newblock \emph{Advances in Neural Information Processing Systems}, pages
  7576--7586, 2018.

\bibitem[Balandat et~al.(2020)Balandat, Karrer, Jiang, Daulton, Letham, Wilson,
  and Bakshy]{balandat2020botorch}
Maximilian Balandat, Brian Karrer, Daniel~R. Jiang, Samuel Daulton, Benjamin
  Letham, Andrew~Gordon Wilson, and Eytan Bakshy.
\newblock {BoTorch: A Framework for Efficient Monte-Carlo Bayesian
  Optimization}.
\newblock In \emph{Advances in Neural Information Processing Systems 33}, 2020.

\bibitem[Bajaj et~al.(2018)Bajaj, Iyer, and Hasan]{bajaj2018trust}
Ishan Bajaj, Shachit~S Iyer, and MM~Faruque Hasan.
\newblock A trust region-based two phase algorithm for constrained black-box
  and grey-box optimization with infeasible initial point.
\newblock \emph{Computers \& Chemical Engineering}, 116:\penalty0 306--321,
  2018.

\bibitem[Kazi et~al.(2021)Kazi, Short, and Biegler]{kazi2021trust}
Saif~R Kazi, Michael Short, and Lorenz~T Biegler.
\newblock A trust region framework for heat exchanger network synthesis with
  detailed individual heat exchanger designs.
\newblock \emph{Computers \& chemical engineering}, 153:\penalty0 107447, 2021.

\bibitem[Kazi et~al.(2022)Kazi, De~Mel, and Short]{kazi2022new}
Saif~R Kazi, Ishanki~A De~Mel, and Michael Short.
\newblock A new trust-region approach for optimization of multi-period heat
  exchanger networks with detailed shell-and-tube heat exchanger designs.
\newblock In \emph{Computer Aided Chemical Engineering}, volume~49, pages
  241--246. Elsevier, 2022.

\bibitem[Wilson et~al.(2020)Wilson, Borovitskiy, Terenin, Mostowsky, and
  Deisenroth]{wilson2020efficiently}
James Wilson, Viacheslav Borovitskiy, Alexander Terenin, Peter Mostowsky, and
  Marc Deisenroth.
\newblock {Efficiently Sampling Functions from Gaussian Process Posteriors}.
\newblock In \emph{International Conference on Machine Learning}, pages
  10292--10302, 13-18 Jul 2020.

\bibitem[Kingma and Ba(2014)]{kingma2014adam}
Diederik Kingma and Jimmy Ba.
\newblock {Adam: A Method for Stochastic Optimization}.
\newblock \emph{International Conference on Learning Representations}, 12 2014.

\end{thebibliography}






\newpage
\appendix
\onecolumn

\section{Notation Table}

\textbf{General Notation} \\
\begin{center}
\begin{tabular}{|p{8em}  C{39em}|}
     $\mathcal{X}$ & Input space / Search space \\
     $f_*, \ x_*$ & Optimal function value, input at which the optimum is achieved \\
     $t$ & Time-index \\
     $x_i^{(m)}$ & Input to experiment, indexed by $i$, at fidelity $m$ \\
     $T$ & Final time-step \\
     $M$ & Number of different fidelities \\
     $m$ & Fidelity index, must lie in $\{0, 1, ..., M\}$ \\
     $f$ & Latent function we want to optimize  \\
     $D_t$ & Data-set at time $t$ \\
     $\eta^2$ & Variance of observation noise \\ 
     $\mu_0$ & Prior GP mean function \\
     $\mu_t$ & Mean of posterior GP at time $t$ \\
     $k_0$ & Prior GP co-variance function \\
     $(k_t)^2$ & Co-variance function of posterior GP at time $t$ \\
     $\tau_i^{(m)}$ & Time-delay between querying the $i$th experiment (at fidelity $m$) and receiving an observation \\
     $\lambda^{(m)}$ & Batch space of an experiment at fidelity $m$ \\
     $\Lambda^{(m)}$ & Maximum total batch space allowed \\
     $\text{TotalBatchSpace}_t$ & The total batch space of the unobserved queries at time $t$
\end{tabular} \\
\end{center}

\textbf{Multi-fidelity Modelling} \\
\begin{center}
\begin{tabular}{|p{8em}  C{39em}|}
     $f^{(m)}$ & Latent function we used to approximate $f$ at the $m$th fidelity  \\
     $\eta^{(m)}$ & Noise level at the $m$-th fidelity \\
     $D^{(m)}$ & Data-set for the $m$-th fidelity \\
     $\mathcal{B}$ & Matrix of task-dependencies, learnt hyper-parameter of linear multi-task kernel \\
     $k^{(I)}$ & Kernel for measuring similarity among inputs \\
     $k^{(F)}$ & Kernel for measuring similarity among fidelities or tasks \\
     $W$ & Number of latent GPs assumed by the model \\
     $w$ & Index of latent GP
\end{tabular}
\end{center}

\textbf{Bayesian Optimization} \\
\begin{center}
\begin{tabular}{|p{8em}  C{39em} |}
     {$\sigma_t(x)$} & Posterior GP standard deviation at time $t$ and input $x$ \\
     $\sigma^{(m)}_t(x)$ & Posterior GP standard deviation at time $t$, input $x$ and fidelity $m$ \\ 
     $\alpha(x)$ & Acquisition function evaluated at input $x$  \\
     $\alpha^{(m)(x)}$ & Upper-confidence bound generated using the $m$-th fidelity \\
     $\zeta^{(m)}_{MAX}$ & Maximum bias between fidelity $m$ and the target fidelity \\
     $\gamma^{(m)}$ & Threshold used to decide if fidelity $m$ should be queried \\
     $I( \cdot \quad ; \quad \cdot)$  & Information gain \\
     $H(\cdot)$ & Entropy \\
     $\mathcal{Q}_t$ & Batch of points being evaluated at time $t$, but still unobserved \\
     $\psi(x; x_j)$ & Penalizing function centered at $x_j$ \\
     $L$ & Lipschitz constant of $f$ \\
     $B_{\delta}(x)$ & Open ball of radius $\delta$, centered at $x$
\end{tabular}
\end{center}

\begin{center}
\begin{tabular}{|p{8em}  C{39em} |}
     $x_{i, t}^{(m)}$ & Input to the $i$th experiment; $t$ denotes the time at which evaluation started and $m$ the fidelity \\
     $y_{i, t'}^{(m)}$ & Output to the $i$th experiment; $t'$ denotes the time at which evaluation finished and $m$ the fidelity \\
     $\epsilon^{(m)}$ & Noise level for $m$th fidelity \\
     $\lambda^{(m)}$, $\lambda_i$ & Batch space of $m$th fidelity, Batch space of $i$th experiment \\
     $\Lambda$ & Maximum cumulative batch space
\end{tabular}
\end{center}

\section{Implementation Details}

We include implementation details about the methods used and the benchmarks. For full details, we refer any reader to the code, available at \url{https://github.com/jpfolch/MFBoom}.

\subsection{Gaussian Processes}

To implement the Gaussian Processes, we used the GPyTorch \citep{gardner2018gpytorch} and PyTorch \cite{paszke2019pytorch} libraries. When using independent Gaussian Processes, we used an RBF kernel:
\begin{equation}
    k_{RBF}(x_1, x_2) = \theta_0 \exp \left(-\frac{1}{2}(x_1 - x_2)^T\Theta^{-2}(x_1 - x_2) \right) \label{eq: rbf_kernel}
\end{equation}
Where $\theta_0$ is a scaling constant, and $\Theta = $diag$(\ell_1, ..., \ell_d)$, where $\ell_i$ represents the length-scale of the $i$th dimension. For Multi-Task Gaussian Processes, we used the kernel:
\begin{equation*}
    K(x, x') = \sum_{w = 1}^W k^{(I)}_w(x, x') \mathcal{B}_w
\end{equation*}
where each $k_w^{(I)}(x, x')$ is given by an RBF kernel, an in Equation  (\ref{eq: rbf_kernel}), each with \textit{different hyper-parameters}. This choice was made to allow the model to vary the hyper-parameters at each fidelity. We further set the hyper-parameters $W = 2M$ and rank$(\mathcal{B}_w) = M$, where $M$ is the number of fidelities in the problem. This choice was arbitrary, but yielded good results. The kernel hyper-parameters whose value we don't explicitly mention were all learnt by maximizing the marginal log-likelihood \citep{rasmussen2005gps}.

To initialize the hyper-parameters, we trained a Gaussian Process model on $80 \log(d)$ randomly sampled data-points, and maximized the marginal log-likelihood. The data-points were then discarded, and the hyper-parameters from the initial GP were used to initialize the hyper-parameters of the GP used by Bayesian Optimization methods. We then retrained the hyper-parameters every time we obtained twenty new observations of either fidelity. To optimize the marginal log-likelihood, we used ADAM \citep{kingma2014adam} across 75 epochs, with a learning rate of 0.1. 

To avoid any ill-behaviour, we also put a Smoothed Box Prior on each hyper-parameter. For length-scales the box went from $0.025$ to $0.6$, for the output-scale from $0.05$ to $2$, for the noise from $10^{-5}$ to $0.2$, and for the prior mean constant from $-1$ to $1$. All these values were chosen based on the scaling of the objective functions.

\subsection{Methods}

All Bayesian Optimization methods were implemented using GPyTorch \citep{gardner2018gpytorch} and BoTorch \citep{balandat2020botorch}. For the synthetic benchmarks, the acquisition functions were optimized using ADAM \citep{kingma2014adam} with a learning rate of 0.01 for 75 epochs. The optimization procedure was first evaluated on 7500d (independent GP methods), 7500 (multi-task based methods), or 3750 (MF-MES), random inputs, and the gradient optimization was multi-started on the best 10 outputs. The differences in number of points evaluated is based on the computational burden of each method.

\subsubsection{TuRBO}

In contrast, TuRBO \citep{eriksson2019scalable} was optimized by grid search inside the trust region, using a Sobol grid of size $\min\left(5000, \max(2000, 200d )\right)$.

To grow, shrink or move the trust region, we \textit{only considered the high-fidelity observations}.

\subsection{MF-MES}

We created our own implementation of MF-MES (see \cite{takeno2020multi} for notation that follows). We generated $100$ samples of $f_* := \max {f(x)}$ using Thompson Sampling on a Sobol grid of size $7500$. We also used $100$ fantasies of $f_\mathcal{Q}$ to approximate the integral when doing asynchronous Bayesian Optimization.

The numerical integration was done using the Trapezium Rule, with the integration range broken into $500$ intervals. For the integration range, we used $[ -10^{k}, 10^{k} ]$, where $k$ was chosen the be the smallest element from $\{-6, ..., 2 \}$, such that $\eta(10^k) < 10^{-30}$.

Whenever we would detected issues from singular matrices while calculating MF-MES, we added $10^{-30}$ to the diagonal to obtain numerically stable behaviour.

\subsection{Benchmark Details} \label{sec: benchmark_details}

All benchmarks are considered in the hyper-cube $[0, 1]^d$, where $d$ is the problem dimension.

\subsubsection{Synthetic Examples}

We took the synthetic examples to be the same as those analysed in \citet{kandasamy2016multifidelityKbandit}, with slightly different scaling. They are given by:

\textbf{Currin Exponential 2D} \\
\begin{align*}
    f^{(2)}(x) &= \frac{1}{10}\left( 1 - \exp\left(\frac{-1}{2x_2}\right)\right) \left(\frac{2300x_1^3 + 1900 x_1^2+ 2092 x_1 + 60}{100 x_1^3 + 500 x_1^2 + 4 x_1 + 20} \right) \\
    f^{(1)}(x) &= \frac{1}{4}f^{(2)}(x_1 + 0.05, x_2 + 0.05) \frac{1}{4}f^{(2)}(x_1 + 0.05, x_2 - 0.05) + \\ 
    & \qquad + \frac{1}{4}f^{(2)}(x_1 - 0.05, x_2 + 0.05) + \frac{1}{4}f^{(2)}(x_1 - 0.05, x_2 - 0.05)
\end{align*}

\textbf{Bad Currin Exponential 2D} \\

For the highest fidelity, we use the same $f^{(2)}$ as in the normal Currin function. For the low fidelity, we use the negative of $f^{(2)}$, i.e. $f^{(1)}(x) = - f^{(2)}(x)$. \\

\textbf{Park Function 4D} \\
\begin{align*}
    10 f^{(2)} &= \frac{x_1}{2} \left(\sqrt{1 + (x_2 + x_3^2)\frac{x_4}{x_1^2} - 1} - 1 \right) + (x_1 + 3x_4)\exp{(1 + \sin(x_3))} \\
    10 f^{(1)} &= \left(1 + \frac{\sin(x_1)}{10} \right)f^{(2)}(x) - 2x_1^2 + x_2^2 + x_3^2 + 0.5
\end{align*}

\textbf{Borehole Function 8D} \\
\begin{align*}
    f^{(2)}(x) &= \frac{2 \pi x_3 (x_4 - x_6)}{100 \log(x_2 / x_1)\left(1 + 2x_7x_3/(\log(x_2x_1)x_1^2x_8)+ x_3 / x_5\right)} \\
    f^{(1)}(x) &= \frac{5x_3(x_4 - x_6)}{100 \log(x_2/x_1)(1.5 + 2x_7x_3 / (\log(x_2/x_1)x_1^2x_8) + x_3 / x_5)}
\end{align*}

\textbf{Hartmann 3D} \\

The $M^{\text{th}}$ fidelity is given by:
$$
f^{(M)}(x) = \sum_{i=1}^4 \alpha_i^{(m)} \exp\left(-\sum_{j = 1}^3 A_{ij}(x_j - P_{ij})^2 \right)
$$
where: \\
\begin{equation*}
A = 
\begin{pmatrix}
3 & 10 & 30 \\
0.1 & 10 & 35 \\
3  & 10  & 30  \\
0.1 & 10 & 35
\end{pmatrix}
, \qquad
P = 10^{-4}
\begin{pmatrix}
3689 & 1170 & 2673 \\
4699 & 4387 & 7470 \\
1091  & 8732  & 5547  \\
381 & 5743 & 8828
\end{pmatrix}
\end{equation*}
and $\alpha^{(m)} = \alpha + (M-m)\delta$ where $\alpha = (1, 1.2, 3, 3.2)$, and $\delta = (0.01, -0.01, -0.1, 0.1)$. Recall that we set $M = 3$. \\

\textbf{Hartmann 6D} \\

The 6D Hartmann takes the same form as the 3D Hartmann except we use the matrices: \\

\begin{equation*}
A = 
\begin{pmatrix}
10 & 3 & 17 & 3.5 & 1.7 & 8 \\
0.05 & 10 & 17 & 0.1 & 8 & 14 \\
3 & 3.5 & 1.7 & 10 & 17 & 8 \\
17 & 8 & 0.05 & 10 & 0.1 & 14 
\end{pmatrix}
, \qquad
P = 10^{-4}
\begin{pmatrix}
1312 & 1696 & 5569 & 124 & 8283 & 5886 \\
2329 & 4135 & 8307 & 3736 & 1004 & 9991 \\
2348 & 1451 & 3522 & 2883 & 3047 & 6650 \\
4047 & 8828 & 8732 & 5743 & 1091 & 381
\end{pmatrix}
\end{equation*}

\textbf{Ackley 40D} \\

For Ackley, we created the fidelities using the following equations:
\begin{align*}
    6 f^{(2)}(x) &=  20 + \exp\left(-0.2 \sqrt{(\frac{1}{40}\sum_{i = 1}^{40}((9x_i - 4)^2)}\right)  + \exp\left(\frac{1}{40}\sum_{i = 1}^{40}\cos(2\pi(9x_i - 4))\right) - 20 - e^1 \\
    6 f^{(1)}(x) &=  20 + \exp\left(-0.2 \sqrt{(\frac{1}{45}\sum_{i = 1}^{40}((9x_i - 3.8)^2)}\right)  + \exp\left(\frac{1}{43}\sum_{i = 1}^{40}\cos(2\pi(9x_i - 3.8))\right) - 20 - e^1
\end{align*}
 
\subsubsection{Battery}

For the battery example, we considered a combinatorial constraint and an equality constraint. The benchmark is 6-dimensional, but we assume only three dimensions are allowed to be active at every moment, i.e.:
\begin{equation}
    x_i = x_j = x_k = 0 \quad \text{for }i \neq j \neq k
\end{equation}
where $i, j, k \in \{1, 2, ..., 6\}$. We further have the equality constraint:
\begin{equation}
    x_{i_a} + x_{j_a} + x_{k_a} = 1
\end{equation}
where $i_a, j_a, k_a$ represent the \textit{active} problem dimensions. To optimize with these constraints, we create a 2-dimensional Sobol grid, and filter out all points such that $x_0 + x_1 > 1$. We then calculate $x_2 = 1 - x_1 - x_0$. Finally, we expand the grid to all 20 possible combinations of active variables.

Due to these constraints, we do not consider TuRBO (trust region and intersection of feasible region might be too small) or MF-MES (grid size is too large to optimize the acquisition function comfortably) on the Battery benchmark.
\end{document}